\documentclass[fontsize=10pt,
  paper=letter,
  german,USenglish
  ]{scrartcl}
\usepackage{babel}
\makeatletter
\defineshorthand{"-}{\penalty\@M-\futurelet\@let@token\goodhyphen@}
\newcommand*\goodhyphen@{%
  \ifmmode\errmessage{Special hyphenation in math mode}\fi
  \if\noexpand\@let@token\relax\else
  \@tempswafalse
  \catcode`\a=11
  \catcode`\*=12
  \ifcat\@let@token a\@tempswatrue\fi
  \ifcat\@let@token *\@tempswatrue\fi
  \if@tempswa
    {\setbox\z@\hbox{-\@let@token}\setbox\tw@\hbox{-\hskip\z@skip\@let@token}%
    \hskip\dimexpr\wd\z@-\wd\tw@\relax}%
  \fi
  \fi
}

\useshorthands{"}
\usepackage[utf8]{inputenc}
\usepackage{lmodern}
\normalfont
\usepackage[T1]{fontenc}
\areaset{142.5mm}{\dimexpr\topskip+49\baselineskip\relax}
\setkomafont{sectioning}{\bfseries}
\normallineskiplimit\normallineskip 
\usepackage{amsmath}
\usepackage{amsthm}
\usepackage{amsfonts}
\usepackage{ifpdf}
  \ifpdf
    \newcommand*\driver{}
  \else
    \newcommand*\driver{dvipdfmx}
  \fi
\usepackage[%
  pdftitle={Modern hierarchical, agglomerative clustering algorithms},
  pdfauthor={Daniel Müllner},
  pdfsubject={},
  pdfdisplaydoctitle=true,
  pdfduplex=DuplexFlipLongEdge,
  colorlinks=True,
  pdfhighlight=/I,
  pdfborder={0 0 0}, 
  linkbordercolor={1 .8 .8},
  citebordercolor={.5 .9 .5},
  urlbordercolor={.5 .7 1},
  linkcolor={black},
  citecolor={black},
  urlcolor={black},
  pdfpagemode=UseOutlines,
  bookmarksopen=true,
  bookmarksopenlevel=1,
  bookmarksdepth=2,
  breaklinks=true,
  unicode=true,
  \driver
]{hyperref}
\usepackage[all]{hypcap}
\ifpdf\usepackage{hypdestopt}\fi

\addto\extrasUSenglish{%

}
\newcommand*\mynewtheorem[2]{%
  \newtheorem{#1}{#2}
  \expandafter\let\csname c@#1\endcsname=\c@theorem
  \expandafter\def\csname#1autorefname\endcsname{#2}%
}
\renewcommand\thmhead[3]{%
  \thmname{#1}\thmnumber{\@ifnotempty{#1}{ }\@upn{#2}}%
  \thmnote{ {\the\thm@notefont#3}}}
\mynewtheorem{theorem}{Theorem}
\mynewtheorem{lemma}{Lemma}
\mynewtheorem{example}{Example}
\mynewtheorem{cor}{Corollary}
\mynewtheorem{myproposition}{Proposition}
\theoremstyle{definition}

\newtheorem*{definition}{Definition}

\RequirePackage{natbib}
\bibpunct{(}{)}{;}{a}{,}{,}
\usepackage[\driver]{xcolor}
\usepackage[\driver]{graphicx}
\usepackage[arrow,pdf,frame,matrix]{xy}
\usepackage{algpseudocode}
\makeatletter
\newcounter{procedure}
\newcommand*\prochypertarget[1]{%
  \stepcounter{procedure}%
  \Hy@raisedlink{\hyper@anchorstart{procedure.\the\c@procedure}\hyper@anchorend}%
  \def\@currentHref{procedure.\the\c@procedure}%
  \def\@currentlabelname{\textproc{#1}}%
}
\algnewcommand\algorithmicclass{\textbf{class}}
\algnewcommand\algorithmicmethod{\textbf{method}}
\algblockdefx[CLASS]{Class}{EndClass}[1]{%
  \prochypertarget{#1}%
  \algorithmicclass\ \textproc{#1}}{\algorithmicend\ \algorithmicclass}
\algblockdefx[METHOD]{Method}{EndMethod}[2]{%
  \prochypertarget{#1}%
  \algorithmicmethod\ \textproc{#1}\ifthenelse{\equal{#2}{}}{}{(#2)}}%
  {\algorithmicend\ \algorithmicmethod}
\algblockdefx[PROCEDURE]{Procedure}{EndProcedure}%
   [2]{%
   \prochypertarget{#1}%
   \algorithmicprocedure\ \textproc{#1}\ifthenelse{\equal{#2}{}}{}{(#2)}}%
   {\algorithmicend\ \algorithmicprocedure}%

\ifthenelse{\equal{\ALG@noend}{t}}%
  {%
  \algtext*{EndClass}%
  \algtext*{EndMethod}%
  \algtext*{EndProcedure}%
  }{}%
\makeatother
\usepackage{float}

\floatstyle{ruled}
\newfloat{algorithm}{tbp}{alg}
\floatname{algorithm}{Figure}

\makeatletter
\renewcommand\fs@ruled{\def\@fs@cfont{\bfseries}\let\@fs@capt\floatc@ruled
\def\@fs@pre{\hrule height.8pt depth0pt \kern4pt}%
\def\@fs@post{\kern4.5pt\hrule height.8pt\relax}
\def\@fs@mid{\kern3.5pt\hrule\kern4pt}%
\let\@fs@iftopcapt\iftrue}
\makeatother
\newcommand*\centercolon[1]{\vcenter{\mathsurround0pt\hbox{$#1:$}}}
\newcommand*\coloneq{\ensuremath{\mathrel{\mathpalette\centercolon=}}}
\newcommand*\co{\:{:}\;}

\newcommand*\None{\textup{None}}
\DeclareMathOperator\argmin{argmin}
\DeclareMathOperator\length{length}
\newcommand*\parent{\mathit{parent}}
\newcommand*\nextlabel{\mathit{nextlabel}}
\newcommand*\size{\mathit{size}}
\newcommand*\nnghbr{\mathit{n\_nghbr}}
\newcommand*\mindist{\mathit{mindist}}

\newcommand*\amsatop[2]{{\genfrac{}{}{0pt}{1}{#1}{#2}}}
\newcommand\doi[1]{doi: \href{http://dx.doi.org/#1}{#1}}

\newcommand*\lightrulewidth{.04em}

\title{Modern hierarchical, agglomerative clustering algorithms}
\author{\scshape Daniel Müllner}
\date{}

\begin{document}
\maketitle

\begin{abstract}\noindent
This paper presents algorithms for hierarchical, agglomerative clustering which perform most efficiently in the general-purpose setup that is given in modern standard software. Requirements are: (1) the input data is given by pairwise dissimilarities between data points, but extensions to vector data are also discussed (2) the output is a “stepwise dendrogram”, a data structure which is shared by all implementations in current standard software. We present algorithms (old and new) which perform clustering in this setting efficiently, both in an asymptotic worst-case analysis and from a practical point of view. The main contributions of this paper are: (1) We present a new algorithm which is suitable for any distance update scheme and performs significantly better than the existing algorithms. (2) We prove the correctness of two algorithms by Rohlf and Murtagh, which is necessary in each case for different reasons. (3) We give well-founded recommendations for the best current algorithms for the various agglomerative clustering schemes.
\bigskip

\end{abstract}

\section{Introduction}\label{sec:introduction}

Hierarchical, agglomerative clustering is an important and well-established technique in unsupervised machine learning. Agglomerative clustering schemes start from the partition of the data set into singleton nodes and merge step by step the current pair of mutually closest nodes into a new node until there is one final node left, which comprises the entire data set. Various clustering schemes share this procedure as a common definition, but differ in the way in which the measure of inter-cluster dissimilarity is updated after each step. The seven most common methods are termed single, complete, average (UPGMA), weighted (WPGMA, McQuitty), Ward, centroid (UPGMC) and median (WPGMC) linkage \citep[see][Table 4.1]{Everitt}. They are implemented in standard numerical and statistical software such as R \citep{R}, MATLAB \citep{Matlab}, Mathematica \citep{Mathematica}, SciPy \citep{SciPy}.

The stepwise, procedural definition of these clustering methods directly gives a valid but inefficient clustering algorithm. Starting with Gower's and Ross's observation \citep{Gower_Ross} that single linkage clustering is related to the minimum spanning tree of a graph in 1969, several authors have contributed algorithms to reduce the computational complexity of agglomerative clustering, in particular \citet{Sibson1973}, \citet{Rohlf1973}, \citet[page 135]{Anderberg1973}, \citet{Murtagh1984}, \citet[Table 5]{Day_Edelsbrunner1984}.

Even when software packages do not use the inefficient primitive algorithm (as SciPy \citep{SciPy-linkage} and the R \citep{R} methods \texttt{hclust} and \texttt{agnes} do), the author found that implementations largely use suboptimal algorithms rather than improved methods suggested in theoretical work. This paper is to advance the theory further up to a point where the algorithms can be readily used in the standard setting, and in this way bridge the gap between the theoretical advances that have been made and the existing software implementations, which are widely used in science and industry.

The main contributions of this paper are:
\begin{itemize}
\item
We present a new algorithm which is suitable for any distance update scheme and performs significantly better than the existing algorithms for the ``centroid'' and ``median'' clustering schemes.

\item
We prove the correctness of two algorithms, a single linkage algorithm by \citet{Rohlf1973} and Murtagh's nearest-neighbor-chain algorithm \citep[page 86]{Murtagh1985}. These proofs were still missing, and we detail why the two proofs are necessary, each for different reasons.

\item
These three algorithms \citep[together with an alternative by][]{Sibson1973} are the best currently available ones, each for its own subset of agglomerative clustering schemes. We justify this carefully, discussing potential alternatives.
\end{itemize}

The specific class of clustering algorithms which is dealt with in this paper has been characterized by the acronym SAHN (sequential, agglomerative, hierarchic, nonoverlapping methods) by \citet[\textsection\,5.4, 5.5]{Sneath_Sokal}. The procedural definition (which is given in \autoref{alg:primitive} below) is not the only possibility for a SAHN method, but this method together with the seven common distance update schemes listed above is most widely used. The scope of this paper is contained further by practical considerations: We consider methods here which comply to the input and output requirements of the general-purpose clustering functions in modern standard software:
\begin{itemize}
\item The input to the algorithm is the list of $\binom N2$ pairwise dissimilarities between $N$ points. (We mention extensions to vector data in \autoref{sec:vectordata}.)

\item The output is a so called \emph{stepwise dendrogram} (see \autoref{def:stepdendr}), in contrast to laxly specified output structure or weaker notions of (non-stepwise) dendrograms in earlier literature.
\end{itemize}
The first item has always been a distinctive characteristic to previous authors since the input format broadly divides into the \textit{stored matrix approach} \citep[\textsection\,6.2]{Anderberg1973} and the \textit{stored data approach} \citep[\textsection\,6.3]{Anderberg1973}. In contrast, the second condition has not been given attention yet, but we will see that it affects the validity of algorithms.

We do not aim to present and compare all available clustering algorithms but build upon the existing knowledge and present only the algorithms which we found the best for the given purpose. For reviews and overviews we refer to \citet{Rohlf1982}, \citet{Murtagh1983, Murtagh1985}, \citet[\textsection 3.1]{Gordon1987}, \citet[\textsection\,3.2]{Jain_Dubes1988}, \citet[\textsection\,4.2]{Day1996}, \citet{Hansen_Jaumard1997}. Those facts about alternative algorithms which are necessary to complete the discussion and which are not covered in existing reviews are collected in \autoref{sec:alternatives}.

The paper is structured as follows:\nopagebreak

\autoref{sec:defns} contains the definitions for input and output data structures as well as specifications of the distance update formulas and the ``primitive'' clustering algorithm.

\autoref{sec:algorithms} is the main section of this paper. We present and discuss three algorithms: our own ``generic algorithm'', Murtagh's nearest-neighbor-chain algorithm and Rohlf's algorithm based on the minimum spanning tree of a graph. We prove the correctness of these algorithms.

\autoref{sec:performance} discusses the complexity of the algorithms, both as theoretical, asymptotic complexity in \autoref{sec:worstcase} and by use-case performance experiments in \autoref{sec:use-case}. We conclude this section by recommendations on which algorithm is the best one for each distance update scheme, based on the preceding analysis.

\autoref{sec:alternatives} discusses alternative algorithms, and \autoref{sec:vectordata} gives a short outlook on extending the context of this paper to vector data instead of dissimilarity input. The paper ends with a brief conclusion in \autoref{sec:conclusion}.

The algorithms in this paper have been implemented in C++ by the author and are available with interfaces to the statistical software R and the programming language Python \citep{Python}. This implementation is presented elsewhere \citep{Muellner_fastcluster}.

\section{Data structures and the algorithmic definition of SAHN clustering methods}\label{sec:defns}

In this section, we recall the common algorithmic (procedural) definition of the SAHN clustering methods which demarcate the scope of this paper. Before we do so, we concretize the setting further by specifying the input and output data structures for the clustering methods. Especially the output data structure has not been specifically considered in earlier works, but nowadays there is a \textit{de facto} standard given by the shared conventions in the most widely used software. Hence, we adopt the setting from practice and specialize our theoretical consideration to the modern standard of the stepwise dendrogram. Later, \autoref{sec:alternatives} contains an example of how the choice of the output data structure affects the result which algorithms are suitable and/or most efficient.

\subsection{Input data structure}\label{sec:input}

The input to the hierarchical clustering algorithms in this paper is always a finite set together with a dissimilarity index \citep[see][\textsection~2.1]{Hansen_Jaumard1997}.

\begin{definition}
A \emph{dissimilarity index} on a set $S$ is a map $d\co S\times S\to [0,\infty)$ which is reflexive and symmetric, i.e.\ we have $d(x,x)=0$ and $d(x,y) = d(y,x)$ for all $x,y\in S$.
\end{definition}

A metric on $S$ is certainly a dissimilarity index. In the scope of this paper, we call the values of $d$ \emph{distances} in a synonymous manner to \emph{dissimilarities}, even though they are not required to fulfill the triangle inequalities, and dissimilarities between different elements may be zero.

If the set $S$ has $N$ elements, a dissimilarity index is given by the $\binom N2$ pairwise dissimilarities. Hence, the input size to the clustering algorithms is $\Theta(N^2)$. Once the primitive clustering algorithm is specified in \autoref{sec:primitive}, it is easy to see that the hierarchical clustering schemes are sensitive to each input value. More precisely, for every input size $N$ and for every index pair $i\neq j$, there are two dissimilarities which differ only at position $(i,j)$ and which produce different output. Hence, all input values must be processed by a clustering algorithm, and therefore the run-time is bounded below by $\Omega(N^2)$.

This bound applies to the general setting when the input is a dissimilarity index. In a different setting, the input could also be given as $N$ points in a normed vector space of dimension $D$ \citep[the ``stored data approach'',][\textsection 6.3]{Anderberg1973}. This results in an input size of $\Theta(ND)$, so that the lower bound does not apply for clustering of vector data. See \autoref{sec:vectordata} for a discussion to which extent the algorithms in this paper can be used in the ``stored data approach''.

\subsection{Output data structures}\label{sec:output}

The output of a hierarchical clustering procedure is traditionally a \emph{dendrogram}. The term ``dendrogram'' has been used with three different meanings: a mathematical object, a data structure and a graphical representation of the former two. In the course of this section, we define a data structure and call it \emph{stepwise dendrogram}. A graphical representation may be drawn from the data in one of several existing fashions. The graphical representation might lose information (e.g.\ when two merges happen at the same dissimilarity value), and at the same time contain extra information which is not contained in the data itself (like a linear order of the leaves).

In the older literature, e.g.\ \citet{Sibson1973}, a dendrogram (this time, as a mathematical object) is rigorously defined as a piecewise constant, right-continuous map $D\co [0,\infty)\to \mathcal P(S)$, where $\mathcal P(S)$ denotes the partitions of $S$, such that
\begin{itemize}
\item $D(s)$ is always coarser than or equal to $D(t)$ for $s>t$,
\item $D(s)$ eventually becomes the one-set partition $\{S\}$ for large $s$.
\end{itemize}

A dendrogram in this sense with the additional condition that $D(0)$ is the singleton partition is in one-to-one correspondence to an \textit{ultrametric} on $S$ \citep[\textsection~I]{Johnson1967}. The ultrametric distance between $x$ and $y$ is given by $\mu(x,y)=\min\{s\geq 0\mid x\sim y \text{ in } D(s)\}$. Conversely, the partition at level $s\geq 0$ in the dendrogram is given by the equivalence relation $x\sim y\Leftrightarrow \mu(x,y)\leq s$. Sibson's ``pointer representation'' and ``packed representation'' \citep[\textsection~4]{Sibson1973} are examples of data structures which allow the compact representation of a dendrogram or ultrametric.

In the current software, however, the output of a hierarchical clustering procedure is a different data structure which conveys potentially more information. We call this a \emph{stepwise dendrogram}.

\begin{definition}\label{def:stepdendr}
Given a finite set $S_0$ with cardinality $N=|S_0|$, a \emph{stepwise dendrogram} is a list of $N-1$ triples $(a_i, b_i, \delta_i)$ $(i=0,\ldots,N-2)$ such that $\delta_i\in[0,\infty)$ and $a_i,b_i\in S_i$, where $S_{i+1}$ is recursively defined as $(S_i\setminus\{a_i,b_i\}) \cup n_i$ and $n_i\notin S\setminus\{a_i,b_i\}$ is a label for a new node.
\end{definition}

This has the following interpretation: The set $S_0$ are the initial data points. In each step, $n_i$ is the new node which is formed by joining the nodes $a_i$ and $b_i$ at the distance $\delta_i$. The order of the nodes within each pair $(a_i, b_i)$ does not matter. The procedure contains $N-1$ steps, so that the final state is a single node which contains all $N$ initial nodes.

(The mathematical object behind this data structure is a sequence of $N+1$ distinct, nested partitions from the singleton partition to the one-set partition, together with a nonnegative real number for each partition. We do not need this abstract point of view here, though.)

The identity of the new labels $n_i$ is not part of the data structure; instead it is assumed that they are generated according to some rule which is part of the specific data format convention. In view of this, it is customary to label the initial data points and the new nodes by integers. For example, the following schemes are used in software:
\begin{itemize}
\item R convention: $S_0\coloneq (-1,\ldots, -N)$, new nodes: $(1,\ldots, N-1)$
\item SciPy convention: $S_0\coloneq (0,\ldots, N-1)$, new nodes: $(N,\ldots, 2N-2)$
\item MATLAB convention: $S_0\coloneq (1,\ldots, N)$, new nodes: $(N+1,\ldots, 2N-1)$
\end{itemize}

We regard a stepwise dendrogram both as an $((N-1)\times 3)$-matrix or as a list of triples, whichever is more convenient in a given situation.

If the sequence $(\delta_i)$ in \autoref{def:stepdendr} is non-decreasing, one says that the stepwise dendrogram does not contain \emph{inversions}, otherwise it does.

In contrast to the first notion of a dendrogram above, a stepwise dendrogram can take inversions into account, which an ordinary dendrogram cannot. Moreover, if more than two nodes are joined at the same distance, the order of the merging steps does matter in a stepwise dendrogram.
\phantomsection\label{tie data set}%
Consider e.g.\ the following data sets with three points:
\[
 \xy <20pt,0cm>:
  (-.5,1.8) *{(A)},
  (0,0)                        *+<8pt>!U!R(.8){x_1} *=<0pt,0pt>{\bullet}
  \PATH ~={**\dir{--}} '(3,0)  *+<8pt>!U!L(.8){x_2} *=<0pt,0pt>{\bullet} _{3.0}
                  '(1.5,1.322) *+<10pt>!D{x_0} *=<0pt,0pt>{\bullet} _{2.0}
  (0,0) _{2.0}
  \POS
  (6,0); p+<20pt,0cm>:
  (-.5,1.8) *{(B)},
  (0,0)                        *+<8pt>!U!R(.8){x_2} *=<0pt,0pt>{\bullet}
  \PATH ~={**\dir{--}} '(3,0)  *+<8pt>!U!L(.8){x_0} *=<0pt,0pt>{\bullet} _{3.0}
                  '(1.5,1.322) *+<10pt>!D{x_1} *=<0pt,0pt>{\bullet} _{2.0}
  (0,0) _{2.0}
  \POS
  (6,0); p+<20pt,0cm>:
  (-.5,1.8) *{(C)},
  (0,0)                        *+<8pt>!U!R(.8){x_0} *=<0pt,0pt>{\bullet}
  \PATH ~={**\dir{--}} '(3,0)  *+<8pt>!U!L(.8){x_1} *=<0pt,0pt>{\bullet} _{3.0}
                  '(1.5,1.322) *+<10pt>!D{x_2} *=<0pt,0pt>{\bullet} _{2.0}
  (0,0) _{2.0}
 \endxy
\]
The array
\[
\begin{bmatrix}
0,&1,&2.0\\
2,&3,&2.0
\end{bmatrix}
\]
is a valid output (SciPy conventions) for single linkage clustering on the data sets $(A)$ and $(B)$ but not for $(C)$. Even more, there is no stepwise dendrogram which is valid for all three data sets simultaneously. On the other hand, the non-stepwise single linkage dendrogram is the same in all cases:
\[
 D(s) =\begin{cases}
   \{x_0\},\{x_1\},\{x_2\} & \text{if $s<2$}\\
   \{x_0, x_1, x_2\} & \text{if $s\geq 2$}
 \end{cases}
 \qquad
 \text{pictorially:}\quad
 \vcenter{\xy <24pt,0cm>:<0pt, 12pt>::
 (0,0) *+!U{x_0} \ar@{-} '(0,1) '(2,1) (2,0) *+!U{x_2}
 \POS
 (1,0) *+!U{x_1} \ar@{-} (1,1.5)
 \POS
 (2,1) *+!L{\scriptstyle 2.0}
 \endxy}
\]
Hence, a stepwise dendrogram conveys slightly more information than a non-stepwise dendrogram in the case of \emph{ties} (i.e.\ when more than one merging step occurs at a certain distance). This must be taken into account when we check the correctness of the algorithms. Although this complicates the proofs in Sections \ref{sec:MST} and \ref{sec:NN-chain} and takes away from the simplicity of the underlying ideas, it is not a matter of hairsplitting: E.g.\ Sibson's SLINK algorithm \citep{Sibson1973} for single linkage clustering works flawlessly if all distances are distinct but produces the same output on all data sets $(A)$, $(B)$ and $(C)$. Hence, the output cannot be converted into a stepwise dendrogram. See \autoref{sec:alternatives} for further details.

\subsection{Node labels}

The node labels $n_i$ in a stepwise dendrogram may be chosen as unique integers according to one of the schemes described in the last section. In an implementation, when the dissimilarities are stored in a large array in memory, it is preferable if each node label $n_i$ for the joined cluster reuses one of the indices $a_i, b_i$ of its constituents, so that the dissimilarities can be updated in-place. Since the clusters after each row in the dendrogram form a partition of the initial set $S_0$, we can identify each cluster not only by its label but also by one of its members. Hence, if the new node label $n_i$ is chosen among $a_i,b_i$, this is sufficient to reconstruct the partition at every stage of the clustering process, and labels can be converted to any other convention in a postprocessing step. Generating unique labels from cluster representatives takes only $\Theta(N)$ time and memory with a suitable union-find data structure. See \autoref{sec:NN-chain} and \autoref{fig:labeling} for details.

\subsection{The primitive clustering algorithm}\label{sec:primitive}

The solution that we expect from a hierarchical clustering algorithm is defined procedurally. All algorithms in this paper are measured against the primitive algorithm in \autoref{alg:primitive}. We state it in a detailed form to point out exactly which information about the clusters is maintained: the pairwise dissimilarities and the number of elements in each cluster.

\begin{algorithm}
\begin{algorithmic}[1]
\Procedure{Primitive\_clustering}{$S, d$}\label{primitive_clustering}
\Comment{$S$: node labels, $d$: pairwise dissimilarities}
\State{$N\gets |S|$}
\Comment{Number of input nodes}
\State{$L\gets[\,]$}
\Comment{Output list}
\State{$\size[x]\gets 1$ for all $x\in S$}
\For{$i\gets 0,\ldots,N-2$}
\State{$(a,b)\gets \argmin_{(S\times S)\setminus \Delta} d$}\label{argmin}
\State{Append $(a,b,d[a,b])$ to $L$.}
\State{$S\gets S\setminus\{a,b\}$}
\State Create a new node label $n\notin S$.
\State{Update $d$ with the information
\[
\advance\displayindent2\dimexpr\algorithmicindent\relax
d[n,x]=d[x,n]=\nameref{formula}(d[a,x], d[b,x],d[a,b], \size[a], \size[b], \size[x])
\prochypertarget{Formula}
\label{formula1}
\]
\hskip2\dimexpr\algorithmicindent\relax
for all $x\in S$.}
\State{$\size[n]\gets\size[a]+\size[b]$}
\State $S\gets S\cup\{n\}$
\EndFor
\State{\Return{$L$}}
\Comment{the stepwise dendrogram, an $((N-1)\times 3)$-matrix}
\EndProcedure
\end{algorithmic}
(As usual, $\Delta$ denotes the diagonal in the Cartesian product $S\times S$.)
\caption{Algorithmic definition of a hierarchical clustering scheme.}
\label{alg:primitive}
\end{algorithm}

The function \nameref{formula1} in \autoref{formula1} is the distance update formula, which returns the distance from a node $x$ to the newly formed node $a \cup b$ in terms of the dissimilarities between clusters $a$, $b$ and $x$ and their sizes. The table in \autoref{fig:formula} lists the formulas for the common distance update methods.

\begin{algorithm}
\caption{Agglomerative clustering schemes.}\label{fig:formula}
\prochypertarget{Formula}
\label{formula}
{\openup6pt
\halign to \textwidth
{%
#\tabskip=\fill\qquad&\hfill$\displaystyle#$\hfill\qquad&\hfill$\displaystyle#$\hfill\tabskip=0pt\cr
Name&\vtop{\normalbaselines\hbox{Distance update formula}\hbox{$\textproc{Formula}$ for $d(I\cup J, K)$}}&\omit{\hskip-1em\hfill\vtop{\normalbaselines\hbox{Cluster dissimilarity}\hbox{between clusters $A$ and $B$}}}\cr
\noalign{\vskip3pt\hrule height.4pt\vskip6pt}
single   & \min(d(I,K),d(J,K))
  & \min_{a\in A, b\in B}d[a,b]\cr
complete & \max(d(I,K),d(J,K))
  & \max_{a\in A, b\in B}d[a,b]\cr
average  & \frac{n_Id(I,K)+n_Jd(J,K)}{n_I+n_J}
  & \frac1{|A||B|}\sum_{a\in A}\sum_{b\in B}d[a,b]\cr
weighted & \frac{d(I,K)+d(J,K)}{2}\cr
Ward     & \hskip-2em\sqrt{\frac{(n_I+n_K)d(I,K)+(n_J+n_K)d(J,K)-n_K d(I,J)}{n_I+n_J+n_K}}\hskip-.5em
  & \sqrt{\frac{2|A||B|}{|A|+|B|}}\cdot\|\vec c_A-\vec c_B\|_2\cr
centroid & \sqrt{\frac{n_Id(I,K)+n_Jd(J,K)}{n_I+n_J}-\frac{n_In_Jd(I,J)}{(n_I+n_J)^2}}
  & \|\vec c_A-\vec c_B\|_2\cr
median   & \sqrt{\frac{d(I,K)}2+\frac {d(J,K)}2-\frac{d(I,J)}4}
  & \|\vec w_A-\vec w_B\|_2\cr
}}
\bigskip

\begin{list}{}{\leftmargin0pt\rightmargin\leftmargin}
\item
Legend: Let $I,J$ be two clusters joined into a new cluster, and let $K$ be any other cluster. Denote by $n_I$, $n_J$ and $n_K$ the sizes of (i.e.\ number of elements in) clusters $I, J, K$, respectively.

\item
The update formulas for the “Ward”, “centroid” and “median” methods assume that the input points are given as vectors in Euclidean space with the Euclidean distance as dissimilarity measure. The expression $\vec c_X$ denotes the centroid of a cluster $X$. The point $\vec w_X$ is defined iteratively and depends on the clustering steps: If the cluster $L$ is formed by joining $I$ and $J$, we define $\vec w_L$ as the midpoint $\frac12(\vec w_I+\vec w_J)$.

\item
All these formulas can be subsumed (for squared Euclidean distances in the three latter cases) under a single formula
\[
   d(I\cup J,K) \coloneq \alpha_I d(I,K) + \alpha_J d(J,K) + \beta d(I,J) + \gamma |d(I,K) - d(J,K)|,
\]
where the coefficients $\alpha_I,\alpha_J,\beta$ may depend on the number of elements in the clusters $I$, $J$ and $K$. For example, $\alpha_I=\alpha_J=\frac12$, $\beta=0$, $\gamma=-\frac12$ gives the single linkage formula. All clustering methods which use this formula are combined under the name ``flexible'' in this paper, as introduced by \citet{LanceWilliams}.


\item
References: \citet{LanceWilliams}, \citet[\textsection 5.5.1]{Kaufman_Rousseeuw}
\end{list}
\vskip-\lastskip
\end{algorithm}

For five of the seven formulas, the distance between clusters does not depend on the order which the clusters were formed by merging. In this case, we also state closed, non-iterative formulas for the cluster dissimilarities in the third row in \autoref{fig:formula}. The distances in the “weighted” and the “median” update scheme depend on the order, so we cannot give non-iterative formulas.

The “centroid” and “median” formulas can produce inversions in the stepwise dendrograms; the other five methods cannot. This can be checked easily: The sequence of dissimilarities at which clusters are merged in \autoref{alg:primitive} cannot decrease if the following condition is fulfilled for all disjoint subsets $I,J,K\subset S_0$:
\[
  d(I,J) \leq \min\{d(I,K), d(J,K)\}\quad\Rightarrow\quad  d(I,J)\leq d(I\cup J,K)
\]

On the other hand, configurations with inversion in the ``centroid'' and ``median'' schemes can be easily produced, e.g.\ three points near the vertices of an equilateral triangle in $\mathbb R^2$.

The primitive algorithm takes $\Theta(N^3)$ time since in the $i$-th iteration of $N-1$ in total, all $\binom{N-1-i}{2}\in \Theta(i^2)$ pairwise distances between the $N-i$ nodes in $S$ are searched.

Note that the stepwise dendrogram from a clustering problem $(S,d)$ is not always uniquely defined, since the minimum in \autoref{argmin} of the algorithm might be attained for several index pairs. We consider every possible output of \nameref{primitive_clustering} under any choices of minima as a valid output.

\section{Algorithms}\label{sec:algorithms}

In the main section of this paper, we present three algorithms which are the most efficient ones for the task of SAHN clustering with the stored matrix approach. Two of the algorithms were described previously: The nearest-neighbor chain (``NN-chain'') algorithm by \citet[page 86]{Murtagh1985}, and an algorithm by \citet{Rohlf1973}, which we call ``MST-algorithm'' here since it is based on Prim's algorithm for the minimum spanning tree of a graph. Both algorithms were presented by the respective authors, but for different reasons each one still lacks a correctness proof. Sections \ref{sec:NN-chain} and \ref{sec:MST} state the algorithms in a way which is suitable for modern standard input and output structures and supply the proofs of correctness.

The third algorithm in \autoref{sec:generic} is a new development based on Anderberg's idea to maintain a list of nearest neighbors for each node \citet[pages 135--136]{Anderberg1973}. While we do not show that the worst-case behavior of our algorithm is better than the $O(N^3)$ worst-case complexity of Anderberg's algorithm, the new algorithm is for all inputs at least equally fast, and we show by experiments in \autoref{sec:use-case} that the new algorithm is considerably faster in practice since it cures Anderberg's algorithm from its worst-case behavior at random input.

As we saw in the last section, the solution to a hierarchical clustering task does not have a simple, self-contained specification but is defined as the outcome of the “primitive” clustering algorithm. The situation is complicated by the fact that the primitive clustering algorithm itself is not completely specified: if a minimum inside the algorithm is attained at more than one place, a choice must be made. We do not require that ties are broken in a specific way; instead we consider any output of the primitive algorithm under any choices as a valid solution. Each of the “advanced” algorithms is considered correct if it always returns one of the possible outputs of the primitive algorithm.

\subsection{The generic clustering algorithm}\label{sec:generic}

The most generally usable algorithm is described in this section. We call it \nameref{alg:generic_linkage} since it can be used with any distance update formula. It is the only algorithm among the three in this paper which can deal with inversions in the dendrogram. Consequentially, the “centroid” and “median” methods must use this algorithm.

The algorithm is presented in \autoref{fig:generic_linkage}. It is a sophistication of the primitive clustering algorithm and of Anderberg's approach \citep[pages 135--136]{Anderberg1973}. Briefly, candidates for nearest neighbors of clusters are cached in a priority queue to speed up the repeated minimum searches in line \ref{argmin} of \nameref{primitive_clustering}.

\begin{algorithm}
\begin{algorithmic}[1]
\Procedure{Generic\_\hskip0pt linkage}{$N, d$}\label{alg:generic_linkage}
\Comment{$N$: input size, $d$: pairwise dissimilarities}
\State{$S\gets(0,\ldots,N-1)$}
\State{$L\gets[\,]$}
\Comment{Output list}
\State{$\size[x]\gets 1$ for all $x\in S$}
\For{$x$ in $S\setminus\{N-1\}$}\label{minsearch:start}
\Comment{Generate the list of nearest neighbors.}
\State{$\nnghbr[x]\gets \argmin_{y>x}d[x,y]$}\label{generic:argmin}
\State{$\mindist[x]\gets d[x,\nnghbr[x]]$}
\EndFor\label{minsearch:end}
\State{$Q\gets{}$(priority queue of indices in $S\setminus\{N-1\}$, keys are in $\mindist$)}\label{generateQ}
\For{$i\gets 1,\ldots,N-1$}\label{main:start}
\Comment{Main loop.}
\State{$a\gets\text{(minimal element of $Q$)}$}\label{currentmin:start}
\State{$b\gets\nnghbr[a]$}
\State{$\delta\gets\mindist[a]$}
\While{$\delta\neq d[a,b]$}\label{while:start}
\Comment{Recalculation of nearest neighbors, if necessary.}
\State{$\nnghbr[a]\gets \argmin_{x>a}d[a,x]$}\label{line:bottleneck}
\State{Update $\mindist$ and $Q$ with $(a, d[a,\nnghbr[a]])$}
\State{$a\gets\text{(minimal element of $Q$)}$}
\State{$b\gets\nnghbr[a]$}
\State{$\delta\gets\mindist[a]$}
\EndWhile\label{currentmin:end}\label{while:end}
\State{Remove the minimal element $a$ from $Q$.}
\State{Append $(a,b,\delta)$ to $L$.}\label{normal:start}
\Comment{Merge the pairs of nearest nodes.}
\State{$\size[b]\gets\size[a]+\size[b]$}
\Comment{Re-use $b$ as the index for the new node.}
\State{$S\gets S\setminus\{a\}$}
\For{$x$ in $S\setminus\{b\}$}
\Comment{Update the distance matrix.}
  \State{$d[x,b]\gets d[b,x] \gets \nameref{formula}(d[a,x], d[b,x], d[a,b], \size[a], \size[b], \size[x])$}
\EndFor\label{normal:end}
\For{$x$ in $S$ such that $x<a$}\label{no_a:start}\label{dispose:start}
\Comment{Update \emph{candidates} for nearest neighbors.}
    \If{$\nnghbr[x]=a$}
\Comment{Deferred search; no nearest}
      \State{$\nnghbr[x]\gets b$}
\Comment{neighbors are searched here.}
    \EndIf
\EndFor\label{no_a:end}
\For{$x$ in $S$ such that $x<b$}
    \If{$d[x,b]<\mindist[x]$}\label{newmin:start}
      \State{$\nnghbr[x]\gets b$}
      \State{Update $\mindist$ and $Q$ with $(x, d[x,b])$}\label{line:update queue}
\Comment{Preserve a lower bound.}
    \EndIf\label{newmin:end}
\EndFor\label{dispose:end}
\State{$\nnghbr[b]\gets \argmin_{x>b}d[b,x]$}\label{newnodemin:start}
\State{Update $\mindist$ and $Q$ with $(b, d[b,\nnghbr[b]])$}\label{newnodemin:end}
\EndFor\label{main:end}
\State{\Return{$L$}}
\Comment{The stepwise dendrogram, an $((N-1)\times 3)$-matrix.}
\EndProcedure
\end{algorithmic}
\caption{The generic clustering algorithm.}\label{fig:generic_linkage}
\end{algorithm}

For the pseudocode in \autoref{fig:generic_linkage}, we assume that the set $S$ are integer indices from $0$ to $N-1$. This is the way in which it may be done in an implementation, and it makes the description easier than for an abstract index set $S$. In particular, we rely on an order of the index set (see e.g.\ line \ref{generic:argmin}: the index ranges over all $y>x$).

There are two levels of sophistication from the primitive clustering algorithm to our generic clustering algorithm. In a first step, one can maintain a list of nearest neighbors for each cluster. For the sake of speed, it is enough to search for the nearest neighbors of a node $x$ only among the nodes with higher index $y>x$. Since the dissimilarity index is symmetric, this list will still contain a pair of closest nodes. The list of nearest neighbors speeds up the global minimum search in the $i$-th step from $\binom {N-i-1}2$ comparisons to $N-i-1$ comparisons at the beginning of each iteration. However, the list of nearest neighbors must be maintained:\nolinebreak[1] if the nearest neighbor of a node $x$ is one of the clusters $a,b$ which are joined, then it is sometimes necessary to search again for the nearest neighbor of $x$ among all nodes $y>x$. Altogether, this reduces the best-case complexity of the clustering algorithm from $\Theta(N^3)$ to $\Theta(N^2)$, while the worst case complexity remains $O(N^3)$. This is the method that Anderberg suggested in \citep[pages 135--136]{Anderberg1973}.

On a second level, one can try to avoid or delay the nearest neighbor searches as long as possible. Here is what the algorithm \nameref{alg:generic_linkage} does: It maintains a list $\nnghbr$ of \textbf{candidates} for nearest neighbors, together with a list $\mindist$ of \textbf{lower bounds} for the distance to the true nearest neighbor. If the distance $d[x,\nnghbr[x]]$ is equal to $\mindist[x]$, we know that we have the true nearest neighbor, since we found a realization of the lower bound; otherwise the algorithm must search for the nearest neighbor of $x$ again.

To further speed up the minimum searches, we also make the array $\mindist$ into a priority queue, so that the current minimum can be found quickly. We require a priority queue $Q$ with a minimal set of methods as in the list below. This can be implemented conveniently by a binary heap \citep[see][Chapter 6]{CLRM}. We state the complexity of each operation by the complexity for a binary heap.
\begin{itemize}
\item $\textproc{Queue}(v)$: Generate a new queue from a vector $v$ of length $|v|=N$. Return: an object $Q$. Complexity: $O(N)$.

\item $\textproc{Q.Argmin}$: Return the index to a minimal value of $v$. Complexity: $O(1)$.

\item $\textproc{Q.Remove\_Min}$: Remove the minimal element from the queue. Complexity: $O(\log N)$.

\item $\textproc{Q.Update}(i,x)$: Assign $v[i]\gets x$ and update the queue accordingly. Complexity: $O(\log N)$.
\end{itemize}

We can now describe the \nameref{alg:generic_linkage} algorithm step by step: Lines \ref{minsearch:start} to \ref{minsearch:end} search the nearest neighbor and the closest distance for each point $x$ among all points $y>x$. This takes $O(N^2)$ time. In \autoref{generateQ}, we generate a priority queue from the list of nearest neighbors and minimal distances.

The main loop is from \autoref{main:start} to the end of the algorithm. In each step, the list $L$ for a stepwise dendrogram is extended by one row, in the same way as the primitive clustering algorithm does.

Lines \ref{currentmin:start} to \ref{currentmin:end} find a current pair of closest nodes. A candidate for this is the minimal index in the queue (assigned to $a$), and its candidate for the nearest neighbor $b\coloneq \nnghbr[a]$. If the lower bound $\mindist[a]$ is equal to the actual dissimilarity $d[a,b]$, then we are sure that we have a pair of closest nodes and their distance. Otherwise, the candidates for the nearest neighbor and the minimal distance are not the true values, and we find the true values for $\nnghbr[a]$ and $\mindist[a]$ in \autoref{line:bottleneck} in $O(N)$ time. We repeat this process and extract the minimum among all lower bounds from the queue until we find a valid minimal entry in the queue and therefore the actual closest pair of points.

This procedure is the performance bottleneck of the algorithm. The algorithm might be forced to update the nearest neighbor $O(N)$ times with an effort of $O(N)$ for each of the $O(N)$ iterations, so the worst-case performance is bounded by $O(N^3)$. In practice, the inner loop from lines \ref{while:start} to \ref{while:end} is executed less often, which results in faster performance.

Lines \ref{normal:start} to \ref{normal:end} are nearly the same as in the primitive clustering algorithm. The only difference is that we specialize from an arbitrary label for the new node to re-using the index $b$ for the joined node. The index $a$ becomes invalid, and we replace any nearest-neighbor reference to $a$ by a reference to the new cluster $b$ in lines \ref{no_a:start} to \ref{no_a:end}. Note that the array $\nnghbr$ contains only candidates for the nearest neighbors, so we could have written any valid index here; however, for the single linkage method, it makes sense to choose $b$: if the nearest neighbor of a node was at index $a$, it is now at $b$, which represents the join $a\cup b$.

The remaining code in the main loop ensures that the array $\nnghbr$ still contains lower bounds on the distances to the nearest neighbors. If the distance from the new cluster $x$ to a cluster $b<x$ is smaller than the old bound for $b$, we record the new smallest distance and the new nearest neighbor in lines \ref{newmin:start} to \ref{newmin:end}.

Lines \ref{newnodemin:start} and \ref{newnodemin:end} finally find the nearest neighbor of the new cluster and record it in the arrays $\nnghbr$ and $\mindist$ and the queue $Q$

The main idea behind this approach is that invalidated nearest neighbors are not re-computed immediately. Suppose that the nearest neighbor of a node $x$ is far away from $x$ compared to the global closed pair of nodes. Then it does not matter that we do not know the nearest neighbor of $x$, as long as we have a lower bound on the distance to the nearest neighbor. The candidate for the nearest neighbor might remain invalid, and the true distance might remain unknown for many iterations, until the lower bound for the nearest-neighbor distance has reached the top of the queue $Q$. By then, the set of nodes $S$ might be much smaller since many of them were already merged, and the algorithm might have avoided many unnecessary repeated  nearest-neighbor searches for $x$ in the meantime.


This concludes the discussion of our generic clustering algorithm; for the performance see \autoref{sec:performance}. Our explanation of how the minimum search is improved also proves the correctness of the algorithm: Indeed, in the same way as the primitive algorithm does, the \nameref{alg:generic_linkage} algorithm finds a pair of globally closest nodes in each iteration. Hence the output is always the same as from the primitive algorithm (or more precisely: one of several valid possibilities if the closest pair of nodes is not unique in some iteration).

\subsection{The nearest-neighbor-chain algorithm}\label{sec:NN-chain}

In this section, we present and prove correctness of the nearest-neighbor-chain algorithm (shortly: NN-chain algorithm), which was described by \citet[page 86]{Murtagh1985}. This algorithm can be used for the “single”, “complete”, “average”, “weighted” and “Ward” methods.

The NN-chain algorithm is presented in \autoref{fig:NN-chain} as \nameref{alg:NN-chain}. It consists of the core algorithm \nameref{alg:NN-chain-core} and two postprocessing steps. Because of the postprocessing, we call the output of \nameref{alg:NN-chain-core} an \emph{unsorted dendrogram}. The unsorted dendrogram must first be sorted row-wise, with the dissimilarities in the third column as the sorting key. In order to correctly deal with merging steps which happen at the same dissimilarity, it is crucial that a \textit{stable} sorting algorithm is employed, i.e.\ one which preserves the relative order of elements with equal sorting keys. At this point, the first two columns of the output array $L$ contain the label of a member of the respective cluster, but not the unique label of the node itself. The second postprocessing step is to generate correct node labels from cluster representatives. This can be done in $\Theta(N)$ time with a union-find data structure. Since this is a standard technique, we do not discuss it here but state an algorithm in \autoref{fig:labeling} for the sake of completeness. It generates integer node labels according to the convention in SciPy but can easily be adapted to follow any convention.

We prove the correctness of the NN-chain algorithm in this paper for two reasons:
\begin{itemize}
\item We make sure that the algorithm resolves ties correctly, which was not in the scope of earlier literature.

\item Murtagh claims \citep[page 111]{Murtagh1984}, \citep[bottom of page 86]{Murtagh1985} that the NN-chain algorithm works for any distance update scheme which fulfills a certain “reducibility property”\pagebreak[0]
\begin{equation}\label{eq:reducibility}
 d(I,J) \leq \min\{d(I,K), d(J,K)\} \quad\Rightarrow\quad \min\{d(I,K), d(J,K)\} \leq d(I\cup J,K)
\end{equation}
for all disjoint nodes $I,J,K$ at any stage of the clustering \citep[\textsection\,3]{Murtagh1984}, \citep[\textsection\,3.5]{Murtagh1985}. This is false.\footnote{For example, consider the distance update formula $d(I\cup J, K)\coloneq d(I,K)+d(J,K)+d(I,J)$. This formula fulfills the reducibility condition. Consider the following distance matrix between five points in the first column below. The \nameref{primitive_clustering} algorithm produces the correct stepwise dendrogram in the middle column. However, if the point $A$ is chosen first in \autoref{NN:choice} of \nameref{alg:NN-chain-core}, the algorithm outputs the incorrect dendrogram in the right column.
\[
\begin{array}{c|cccc}
 &B&C&D&E\\
\noalign{\hrule height\lightrulewidth}
A&3&4&6&15\vrule height2.4ex width0pt\\
B&&5&7&12\\
C&&&1&13\\
D&&&&14
\end{array}
\qquad\qquad
\begin{array}{lll}
(C,&D,&1)\\
(A,&B,&3)\\
(AB,&CD,&27)\\
(ABCD,&E,&85)
\end{array}
\qquad\qquad
\begin{array}{lll}
(C,&D,&1)\\
(A,&B,&3)\\
(CD,&E,&28)\\
(AB,&CDE,&87)
\end{array}
\]
\vskip-\lastskip\hrule height0pt\relax}
We give a correct proof which also shows the limitations of the algorithm. In Murtagh's papers \citep{Murtagh1984, Murtagh1985}, it is not taken into account that the dissimilarity between clusters may depend on the order of clustering steps; on the other hand, it is explicitly said that the algorithm works for the ``weighted'' scheme, in which dissimilarities depend on the order of the steps.
\end{itemize}

Since there is no published proof for the NN-chain algorithm but claims which go beyond what the algorithm can truly do, it is necessary to establish the correctness by a strict proof:

\newcommand*\chain{\mathit{chain}}

\begin{algorithm}
\caption{The nearest-neighbor clustering algorithm.}\label{fig:NN-chain}
\begin{algorithmic}[1]
\medskip
\Procedure{NN-chain-linkage}{$S,d$}\label{alg:NN-chain}
\Comment{$S$: node labels, $d$: pairwise dissimilarities}
\State{$L\gets\nameref{alg:NN-chain-core}(N,d)$}
\State{Stably sort $L$ with respect to the third column.}\label{NN-chain:sort}
\State{$L\gets\nameref{convert}(L)$}
\Comment{Find node labels from cluster representatives.}
\State{\Return{$L$}}
\EndProcedure
\end{algorithmic}
\medskip
\begin{algorithmic}[1]
\Procedure{NN-chain-core}{$S, d$}\label{alg:NN-chain-core}
\Comment{$S$: node labels, $d$: pairwise dissimilarities}
\State{$S\gets(0,\ldots,N-1)$}
\State{$\chain=[\,]$}
\State{$\size[x]\gets 1$ for all $x\in S$}
\While{$|S|>1$}
\If{$\length(\chain)\leq3$}
\State{$a\gets \text{(any element of $S$)}$}\label{NN:choice}
\Comment{E.g.\ $S[0]$}
\State{$\chain\gets[a]$}
\State{$b\gets \text{(any element of $S\setminus\{a\}$)}$}
\Comment{E.g.\ $S[1]$}
\Else
\State{$a\gets\chain[-4]$}
\State{$b\gets\chain[-3]$}
\State{Remove $\chain[-1]$, $\chain[-2]$ and $\chain[-3]$}
\Comment{Cut the tail $(x,y,x)$.}
\EndIf
\Repeat\label{PNN:start}
\State{$c\gets \argmin_{x\neq a}d[x,a]$ with preference for $b$}\label{NN-chain:argmin}
\State{$a,b\gets c,a$}
\State{Append $a$ to $\chain$}
\Until{$\length(\chain)\geq 3$ and $a=\chain[-3]$}\label{PNN:end}
\Comment{$a,b$ are reciprocal}
\State{Append $(a,b, d[a,b])$ to $L$}
\Comment{nearest neighbors.}
\State{Remove $a,b$ from $S$}
\State{$n\gets\text{(new node label)}$}
\State{$\size[n]\gets \size[a]+\size[b]$}
\State{Update $d$ with the information
\[
\advance\displayindent2\dimexpr\algorithmicindent\relax
d[n,x]=d[x,n]=\nameref{formula}(d[a,x], d[b,x],d[a,b], \size[a], \size[b], \size[x])
\prochypertarget{Formula}
\label{formula2}
\]
\hskip2\dimexpr\algorithmicindent\relax
for all $x\in S$.}
\State{$S\gets S\cup\{n\}$}
\EndWhile
\State \Return $L$
\Comment{an unsorted dendrogram}
\EndProcedure
\end{algorithmic}
(We use the Python index notation: $\chain[-2]$ is the second-to-last element in the list $\chain$.)
\end{algorithm}

\begin{algorithm}
\begin{algorithmic}[1]
\Procedure{Label}{$L$}\label{convert}
\State{$L'\gets[\,]$}
\State{$N\gets \text{(number of rows in $L$)}+1$}
\Comment{Number of initial nodes.}
\State{$U\gets \text{new }\nameref{unionfind}(N)$}
\For{$(a,b,\delta)$ in $L$}
\State{Append $(U.\nameref{alg:Efficient-Find}(a), U.\nameref{alg:Efficient-Find}(b),\delta)$ to $L'$}
\State{$U.\nameref{union}(a,b)$}
\EndFor
\State{\Return{$L'$}}
\EndProcedure

\Statex
\Class{Union-Find}\label{unionfind}
\Method{Constructor}{$N$}
\Comment{$N$ is the number of data points.}
\State $\parent \gets \mathrm{new}\ \mathrm{int}[2N-1]$
\State $\parent[0,\ldots,2N-2] \gets \None$
\State $\nextlabel \gets N$
\Comment{SciPy convention: new labels start at $N$}
\EndMethod
\Statex

\Method{Union}{$m,n$}\label{union}
\State $\parent[m] = \nextlabel$
\State $\parent[n] = \nextlabel$
\State $\nextlabel\gets\nextlabel+1$
\Comment{SciPy convention: number new labels consecutively}
\EndMethod
\Statex
\Method{Find}{$n$}\label{find}
\Comment{This works but the search process is not efficient.}
\While {$\parent[n]$ is not $\None$}
\State $n\gets \parent[n]$
\EndWhile
\State\Return $n$
\EndMethod
\Statex

\Method{Efficient-Find}{$n$}\label{alg:Efficient-Find}
\Comment{This speeds up repeated calls.}
  \State $p\gets n$
  \While {$\parent[n]$ is not $\None$}
  \State $n\gets \parent[n]$
  \EndWhile
    \While {$\parent[p]\neq n$}
      \State $p,\parent[p] \gets \parent[p], n$
    \EndWhile
\State\Return $n$
\EndMethod
\EndClass
\end{algorithmic}
\caption{A union-find data structure suited for the output conversion.}\label{fig:labeling}
\end{algorithm}

\begin{theorem}\label{thm:NN-chain}
Fix a distance update formula. For any sequence of merging steps and any four disjoint clusters $I,J,K,L$ resulting from these steps, require two properties from the distance update formula:
\begin{itemize}
\item
It fulfills the reducibility property \eqref{eq:reducibility}.

\item The distance $d(I\cup J, K\cup L)$ is independent of whether $(I,J)$ are merged first and then $(K,L)$ or the other way round.
\end{itemize}
Then the algorithm \nameref{alg:NN-chain} produces valid stepwise dendrograms for the given method.
\end{theorem}

\begin{myproposition}\label{prop:NN-chain}
The “single”, “complete”, “average”, “weighted” and “Ward” distance update formulas fulfill the requirements of \autoref{thm:NN-chain}.
\end{myproposition}

\begin{proof}[Proof of \autoref{thm:NN-chain}]
We prove the theorem by induction in the size of the input set $S$. The induction start is trivial since a dendrogram for a one-point set is empty.

We call two nodes $a,b\in S$ \emph{reciprocal nearest neighbors} \citep[“pairwise nearest neighbors” in the terminology of][]{Murtagh1985} if the distance $d[a,b]$ is minimal among all distances from $a$ to points in $S$, and also minimal among all distances from $b$:
\[
   d[a,b] = \min_{\amsatop{x\in S}{x\neq a}}d[a,x] = \min_{\amsatop{x\in S}{x\neq b}}d[b,x].
\]

Every finite set $S$ with at least two elements has at least one pair of reciprocal nearest neighbors, namely a pair which realizes the global minimum distance.

The list $\chain$ is in the algorithm constructed in a way such that every element is a nearest neighbor of its predecessor. If $\chain$ ends in $[\ldots,b,a,b]$, we know that $a$ and $b$ are reciprocal nearest neighbors. The main idea behind the algorithm is that reciprocal nearest neighbors $a,b$ always contribute a row $(a,b,d[a,b])$ to the stepwise dendrogram, even if they are not discovered in ascending order of dissimilarities.

Lines \ref{PNN:start} to \ref{PNN:end} in \nameref{alg:NN-chain-core} clearly find reciprocal nearest neighbors $(a,b)$ in $S$. One important detail is that the index $b$ is preferred in the $\argmin$ search in \autoref{NN-chain:argmin}, if the minimum is attained at several indices and $b$ realizes the minimum. This can be respected in an implementation with no effort, and it ensures that reciprocal nearest neighbors are indeed found. That is, the list $\chain$ never contains a cycle of length $>2$, and a $\chain=[\ldots,b,a]$ with reciprocal nearest neighbors at the end will always be extended by $b$, never with an element $c\neq b$ which coincidentally has the same distance to $a$.

After \autoref{PNN:end}, the chain ends in $(b,a,b)$. The nodes $a$ and $b$ are then joined, and the internal variables are updated as usual.

We now show that the remaining iterations produce the same output as if the algorithm had started with the set $S'\coloneq (S\setminus\{a,b\})\cup\{n\}$, where $n$ is the new node label and the distance array $d$ and the $\size$ array are updated accordingly.

The only data which could potentially be corrupted is that the list $\chain$ could not contain successive nearest neighbors any more, since the new node $n$ could have become the nearest neighbor of a node in the list.

At the beginning of the next iteration, the last elements $(b,a,b)$ are removed from $\chain$. The list $\chain$ then clearly does not contain $a$ or $b$ at any place any more, since any occurrence of $a$ or $b$ in the list would have led to an earlier pair of reciprocal nearest neighbors, before $(b,a,b)$ was appended to the list. Hence, $\chain$ contains only nodes which really are in $S$. Let $e,f$ be two successive entries in $\chain$, i.e.\ $f$ is a nearest neighbor of $e$. Then we know
\begin{align*}
 d[e,f]&\leq d[e,a] & d[a,b] &\leq d[a,e]\\
 d[e,f]&\leq d[e,b] & d[a,b] &\leq d[b,e]
\end{align*}

Together with the reducibility property \eqref{eq:reducibility} (for $I=a$, $J=b$, $K=e$), this implies
$d[e,f]\leq d[e,n]$. Hence, $f$ is still the nearest neighbor of $e$, which proves our assertion.

We can therefore be sure that the remaining iterations of \nameref{alg:NN-chain-core} produce the same output as if the algorithm would be run freshly on $S'$. By the inductive assumption, this produces a valid stepwise dendrogram for the set $S'$ with $N-1$ nodes. \autoref{prop:nn-step} carries out the remainder of the proof, as it shows that the first line $(a,b,d[a,b])$ of the unsorted dendrogram, when it is sorted into the right place in the dendrogram for the nodes in $S'$, is a valid stepwise dendrogram for the original set $S$ with $N$ nodes.
\end{proof}

\begin{myproposition}\label{prop:nn-step}
Let $(S,d)$ be a set with dissimilarities ($|S|>1$). Fix a distance update formula which fulfills the requirements in \autoref{thm:NN-chain}. Let $a, b$ be two distinct nodes in $S$ which are reciprocal nearest neighbors.

Define $S'$ as $(S\setminus\{a,b\})\cup \{n\}$, where the label $n$ represents the union $a\cup b$. Let $d'$ be the updated dissimilarity matrix for $S'$, according to the chosen formula. Let $L'=((a_i,b_i,\delta_i)_{i=0,\ldots, m})$ be a stepwise dendrogram for $S'$. Let $j$ be the index such that all $\delta_i<d[a,b]$ for all $i<j$ and $\delta_i\geq d[a,b]$ for all $i\geq j$. That is, $j$ is the index where the new row $(a,b,d[a,b])$ should be inserted to preserve the sorting order, giving $d[a,b]$ priority over potentially equal sorting keys. Then the array $L$, which we define as\pagebreak[0]
\[\tabskip=\fill
 \halign to \displaywidth{$\displaystyle#$\cr
 \newcommand*\factor{1}
 \left[\begin{array}{r@{}lll@{}l}
 &a_{0} & b_{0} & \delta_{0}&\\[\factor\jot]
 &\ldots&&\\[\factor\jot]
 &a_{j-1} & b_{j-1} & \delta_{j-1}&\\[\factor\jot]
 \llap{$\rightarrow\hskip2em$}&a & b & d[a,b] &\rlap{\hskip2em$\leftarrow$}\\[\factor\jot]
 &a_{j} & b_{j} & \delta_{j}&\\[\jot]
 &\ldots&&\\[\factor\jot]
 &a_{m} & b_{m} & \delta_{m}&
 \end{array}\right]\cr
 \noalign{\hrule height0pt}\cr
}
\]
is a stepwise dendrogram for $(S,d)$.
\end{myproposition}

\begin{proof}
Since $a$ and $b$ are reciprocal nearest neighbors at the beginning, the reducibility property \eqref{eq:reducibility} guarantees that they stay nearest neighbors after any number of merging steps between other reciprocal nearest neighbors. Hence, the first $j$ steps in a dendrogram for $S$ cannot contain $a$ or $b$, since these steps all happen at merging dissimilarities smaller than $d[a,b]$.
This is the point where we must require that the sorting in \autoref{NN-chain:sort} of \nameref{alg:NN-chain} is stable.

Moreover, the first $j$ rows of $L$ cannot contain a reference to $n$: Again by the reducibility property, dissimilarities between $n$ and any other node are at least as big as $d[a,b]$. Therefore, the first $j$ rows of $L$ are correct for a dendrogram for $S$.

After $j$ steps, we know that no inter-cluster distances in $S\setminus\{a,b\}$ are smaller than $d[a,b]$. Also, $d[a,b]$ is minimal among all distances from $a$ and $b$, so the row $(a,b,d[a,b])$ is a valid next row in $L$.

After this step, we claim that the situation is the same in both settings: The sets $S'$ after $j$ steps and the set $S$ after $j+1$ steps, including the last one merging $a$ and $b$ into a new cluster $n$, are clearly equal as partitions of the original set. It is required to check that also the dissimilarities are the same in both settings. This is where we need the second condition in \autoref{thm:NN-chain}:

The row $(a,b,d[a,b])$ on top of the array $L'$ differs from the dendrogram $L$ by $j$ transpositions, where $(a,b,d[a,b])$ is moved one step downwards. Each transposition happens between two pairs $(a,b)$ and $(a_i,b_i)$, where all four nodes are distinct, as shown above. The dissimilarity from a distinct fifth node $x$ to the join $a\cup b$ does not depend on the merging of $a_i$ and $b_i$ since there is no way in which dissimilarities to $a_i$ and $b_i$ enter the distance update formula $\nameref{formula}(d[a,x], d[b,x],d[a,b], \size[a], \size[b], \size[x])$. The symmetric statement holds for the dissimilarity $d[x,a_i\cup b_i]$. The nodes $a,b,a_i,b_i$ are deleted after the two steps, so dissimilarities like $d[a,a_i\cup b_i]$ can be neglected. The only dissimilarity between active nodes which could be altered by the transposition is $d[a\cup b,a_i\cup b_i]$. It is exactly the second condition in \autoref{thm:NN-chain} that this dissimilarity is independent of the order of merging steps. This finishes the proof of \autoref{thm:NN-chain}.
\end{proof}

We still have to prove that the requirements of \autoref{thm:NN-chain} are fulfilled by the ``single'', ``complete'', ``average'', ``weighted'' and ``Ward'' schemes:

\begin{proof}[Proof of \autoref{prop:NN-chain}]
It is easy and straightforward to check from the table in \autoref{fig:formula} that the distance update schemes in question fulfill the reducibility property. Moreover, the table also conveys that the dissimilarities between clusters in the ``single'', ``complete'' and ``average'' schemes do not depend on the order of the merging steps.

For Ward's scheme, the global dissimilarity expression in the third column in \autoref{fig:formula} applies only if the dissimilarity matrix consists of Euclidean distances between vectors (which is the prevalent setting for Ward's method). For a general argument, note that the global cluster dissimilarity for Ward's method can also be expressed by a slightly more complicated expression:
\[
d(A,B) = \sqrt{\frac1{|A|+|B|}\left(2\sum_{a\in A}\sum_{b\in B}d(a,b)^2-\frac{|B|}{|A|}\sum_{a\in A}\sum_{a'\in A}d(a,a')^2
-\frac{|A|}{|B|}\sum_{b\in B}\sum_{b'\in B}d(b,b')^2\right)}
\]
This formula can be proved inductively from the recursive distance update formula for Ward's method, hence it holds independently of whether the data is Euclidean or not. This proves that the dissimilarities in Ward's scheme are also independent of the order of merging steps.

Dissimilarities in the ``weighted'' scheme, however, do in general depend on the order of merging steps. However, the dissimilarity between joined nodes $I\cup J$ and $K \cup L$ is always the mean dissimilarity $\frac14(d[I,K]+d[I,L]+d[J,K]+d[J,L])$, independent of the order of steps, and this is all that is required for \autoref{prop:NN-chain}.
\end{proof}

\subsection{The single linkage algorithm}\label{sec:MST}

In this section, we present and prove correctness of a fast algorithm for single linkage clustering. \citet{Gower_Ross} observed that a single linkage dendrogram can be obtained from a minimum spanning tree (MST) of the weighted graph which is given by the complete graph on the singleton set $S$ with the dissimilarities as edge weights. The algorithm here was originally described by \citet{Rohlf1973} and is based on Prim's algorithm for the MST \citep[see][\textsection\,23.2]{CLRM}.

The single linkage algorithm \nameref{alg:MST-linkage} is given in \autoref{fig:SL}. In the same way as the NN-chain algorithm, it consists of a core algorithm \nameref{alg:MST-linkage-core} and two postprocessing steps. The output structure of the core algorithm is again an unsorted list of clustering steps with node representatives instead of unique labels. As will be proved, exactly the same postprocessing steps can be used as for the NN-chain algorithm.

\begin{algorithm}
\begin{algorithmic}[1]
\Procedure{MST-linkage}{$S,d$}\label{alg:MST-linkage}
\Comment{$S$: node labels, $d$: pairwise dissimilarities}
\State{$L\gets\nameref{alg:MST-linkage-core}(S,d)$}
\State{Stably sort $L$ with respect to the third column.}\label{SL:sort}
\State{$L\gets\nameref{convert}(L)$}
\Comment{Find node labels from cluster representatives.}
\State{\Return{$L$}}
\EndProcedure
\end{algorithmic}
\bigskip
\begin{algorithmic}[1]
\Procedure{MST-linkage-core}{$S_0, d$}\label{alg:MST-linkage-core}
\Comment{$S_0$: node labels, $d$: pairwise dissimilarities}
\State{$L\gets[\,]$}
\State{$c\gets \text{(any element of $S_0$)}$}
\Comment{$c$: current node}
\State{$D_0[s]\gets\infty$ for $s\in S_0\setminus\{c\}$}
\For{$i$ in $(1,\ldots, |S_0|-1)$}
\State{$S_{i}\gets S_{i-1}\setminus\{c\}$}
\For{$s$ in $S_i$}
\State{$D_i[s]\gets \min\{D_{i-1}[s], d[s,c]\}$}\label{MST:loop1}
\EndFor
\State{$n\gets \argmin_{s\in S_i}D_i[s]$}\label{76:argmin}\label{MST:loop2}
\Comment{$n$: new node}
\State{Append $(c, n, D_i[n])$ to $L$}
\State{$c\gets n$}
\EndFor
\State \Return $L$
\Comment{an unsorted dendrogram}
\EndProcedure
\end{algorithmic}
\caption{The single linkage algorithm.}\label{fig:SL}
\end{algorithm}

Rohlf's algorithm in its original version is a full Prim's algorithm and maintains enough data to generate the MST. He also mentions a possible simplification which does not do enough bookkeeping to generate an MST but enough for single linkage clustering. It is this simplification that is discussed in this paper. We prove the correctness of this algorithm for two reasons:
\begin{itemize}
\item
Since the algorithm \nameref{alg:MST-linkage-core} does not generate enough information to reconstruct a minimum spanning tree, one cannot refer to the short proof of Prim's algorithm in any easy way to establish the correctness of \nameref{alg:MST-linkage}.

\item
Like for the NN-chain algorithm in the last section, it is not clear \textit{a priori} that the algorithm resolves ties correctly. A third algorithm can serve as a warning here (see \autoref{sec:alternatives} for more details): There is an other fast algorithm for single linkage clustering, Sibson's SLINK algorithm \citep{Sibson1973}. More or less by coincidence, all three algorithms \nameref{alg:NN-chain-core}, \nameref{alg:MST-linkage-core} and SLINK generate output which can be processed by exactly the same two steps: sorting followed by \nameref{convert}. In case of the SLINK algorithm this works fine if all dissimilarities are distinct but produces wrong \emph{stepwise} dendrograms in situations when two merging dissimilarities are equal. There is nothing wrong with the SLINK algorithm, however. Sibson supplied a proof for the SLINK algorithm in his paper \citep{Sibson1973}, but it is written for a (non-stepwise) dendrogram as the output structure, not for a stepwise dendrogram. Hence, the additional information which is contained in a stepwise dendrogram in the case of ties is not provided by all, otherwise correct algorithms.
\end{itemize}

This should be taken as a warning that ties demand more from an algorithm and must be explicitly taken into account when we prove the correctness of the \nameref{alg:MST-linkage} algorithm below.

\begin{theorem}\label{thm:SL}
The algorithm \nameref{alg:MST-linkage} yields an output which can also be generated by \nameref{primitive_clustering}.
\end{theorem}

We do not explicitly refer to Prim's algorithm in the following, and we make the proof self-contained, since the algorithm does not collect enough information to construct a minimum spanning tree. There are unmistakable similarities, of course, and the author got most of the ideas for this proof from Prim's algorithm \citep[see][\textsection\,23.2]{CLRM}.

Let us first make two observations about the algorithm \nameref{alg:MST-linkage}.

\begin{enumerate}\renewcommand*\labelenumi{(\alph{enumi})}
\item Starting with the full initial set $S_0$, the algorithm  \nameref{alg:MST-linkage-core} chooses a ``current node'' $c$ in each step and removes it from the current set $S_i$ in every iteration.
Let $S^c_i\coloneq S_0\setminus S_i$ be the complement of the current set $S_i$. Then $D_i[s]$ ($s\in S_i$) is the distance from $S_i^c$ to $s$, i.e.
\[ D_i[s]=\min_{t\in S_i^c}d[s,t]. \]

\item Let $L$ be the output of $\nameref{alg:MST-linkage-core}(S,d)$. The $2i$ entries in the first two columns and the first $i$ rows contain only $i+1$ distinct elements of $S$, since the second entry in one row is the first entry in the next row.
\end{enumerate}

We prove \autoref{thm:SL} by the following stronger variant:
\begin{theorem}\label{stronger}
Let $L$ be the output of $\nameref{alg:MST-linkage-core}(S,d)$. For all $n<|S|$, the first $n$ rows of $L$ are an unsorted single linkage dendrogram for the $n+1$ points of $S$ in this list (see Observation (b)).
\end{theorem}

\begin{proof}
We proceed by induction. After the first iteration, the list $L$ contains one triple $(a_0,b_0,\delta_0)$. $\delta_0=D_1[b_0]$ is clearly the dissimilarity $d[a_0, b_0]$, since the array $D_1$ contains the dissimilarities to $a_0$ after the first iteration (Observation (a)).

Let $(a_0,b_0,\delta_0), \ldots, (a_{n},b_{n},\delta_{n})$ be the first $n+1$ rows of $L$. We sort the rows with a stable sorting algorithm as specified in \nameref{alg:MST-linkage}. We leave the postprocessing step \nameref{convert} out of our scope and work with the representatives $a_i$, $b_i$ for the rest of the proof.

Let $s(0),\ldots, s(n)$ be the stably sorted indices (i.e.\ $\delta_{s(i)}\leq \delta_{s(i+1)}$ for all $i$ and $s(i)<s(i+1)$ if $\delta_{s(i)}=\delta_{s(i+1)}$). Let $k$ be the sorted index of the last row $n$. Altogether, we have a sorted matrix
\[
 \left[\begin{array}{r@{}lll@{}l}
 &a_{s(0)} & b_{s(0)} & \delta_{s(0)}&\\[\jot]
 &\ldots&&\\[\jot]
 &a_{s(k-1)} & b_{s(k-1)} & \delta_{s(k-1)}&\\[\jot]
 \llap{$\rightarrow\hskip2em$}&a_{s(k)} & b_{s(k)} & \delta_{s(k)}&\rlap{$\hskip2em\leftarrow$}\\[\jot]
 &a_{s(k+1)} & b_{s(k+1)} & \delta_{s(k+1)}&\\[\jot]
 &\ldots&&\\[\jot]
 &a_{s(n)} & b_{s(n)} & \delta_{s(n)}&
 \end{array}\right]
 \]

The new row is at the index $k$, i.e.\ $(a_{s(k)},b_{s(k)},\delta_{s(k)})=(a_{n},b_{n},\delta_{n})$. The matrix without the $k$-th row is a valid stepwise, single linkage dendrogram for the points $a_0,\ldots, a_{n}$, by the induction hypothesis. (Recall that $b_i=a_{i+1}$.) Our goal is to show that the matrix with its $k$-th row inserted yields a valid single linkage dendrogram on the points $a_0,\ldots, a_{n}, b_{n}$.

\textbf{First step: rows $0$ to $k-1$.} The distance $\delta_{n}$ is the minimal distance from $b_{n}$ to any of the points $a_0,\ldots, a_{n}$. Therefore, the dendrograms for the sets $S^-\coloneq\{a_0,\ldots, a_{n}\}$ and $S^+\coloneq S^-\cup \{b_{n}\}$ have the same first $k$ steps, when all the inter-cluster distances are smaller than or equal to $\delta_{n}$. (If the distance $\delta_{n}$ occurs more than once, i.e.\ when $\delta_{s(k-1)}=\delta_{n}$, we assume by stable sorting that the node pairs which do not contain $b_{n}$ are chosen first.)

Therefore, the first $k$ rows are a possible output of \nameref{primitive_clustering} in the first $k$ steps. After this step, we have the same partial clusters in $S^+$ as in the smaller data set, plus a singleton $\{b_{n}\}$.

\textbf{Second step: row $k$.} The distance $\delta_{n}$ from $b_{n}$ to some point $a_0,\ldots,a_{n}$ is clearly the smallest inter-cluster distance at this point, since all other inter-cluster distances are at least $\delta_{s(k+1)}$, which is greater than $\delta_{s(k)}=\delta_{n}$. Since the output row is $(a_{n}, b_{n},\delta_{n})$, it remains to check that the distance $\delta_{n}$ is realized as the distance from $b_{n}$ to a point in the cluster of $a_{n}$, i.e.\ that $a_{n}$ is in a cluster with distance $\delta_{n}$ to $b_{n}$.

The clusters mentioned in the last sentence refer to the partition of $S^+$ which is generated by the relations $a_{s(0)}\sim b_{s(0)},\ldots, a_{s(k-1)}\sim b_{s(k-1)}$. Since we have $b_{i}=a_{i+1}$, the partition of $S^+$ consists of contiguous chains in the original order of points $a_0,a_1,\ldots,a_{n},b_{n}$.

The diagram below visualizes a possible partition after $k$ steps.
\[
 \setbox0\hbox{$\xy
 \everyentry={\mathstrut}
 \xymatrix@C=3pt{
 a_0 & a_1 & a_2 & a_3 & a_4 & a_5 &\ldots & \ldots & a_m & \ldots & a_{n} & b_{n}
 }
 \POS
  "1,1"."1,3" *\frm<4pt>{-}
  ,"1,4"  *\frm<4pt>{-}
  ,"1,5"."1,6" *\frm<4pt>{-}
  ,"1,8"."1,11" *\frm<4pt>{-}
  ,"1,12" *\frm<4pt>{-}
 \endxy$}\raise\dp0\box0
\]

In this particular example, the distances $\delta_0$, $\delta_1$ and $\delta_4$ are among the first $k$ smallest, while $\delta_2$, $\delta_3$ and $\delta_5$ come later in the sorted order.

Let $\delta_{n}$ be realized as the distance between $b_{n}$ and $a_m$ for some $m\leq n$. Then the dissimilarities between consecutive points in the sequence $a_m$, $b_m=a_{m+1}$, $b_{m+1}=a_{m+2}$, \ldots, $b_{n-1}=a_{n}$ must be less than or equal to $\delta_{n}$; otherwise $b_{n}$ and the dissimilarity $\delta_{n}=d[b_{n},a_m]$ would have been chosen first over these other dissimilarities in one of the first $k$ steps. Since the dissimilarities of all pairs $(a_i, b_i)$ in this chain are not more than $\delta_{n}$, they are contained in the first $k$ sorted triples. Hence, $a_m$ and $a_{n}$ have been joined into a cluster in the first $k$ steps, and $a_{n}$ is a valid representative of the cluster that also contains $a_m$.

Note that the argument in the last paragraph is the point where we need that the sorting in \autoref{SL:sort} of \nameref{alg:MST-linkage} is stable. Otherwise it could not be guaranteed that $a_m$ and $a_{n}$ have been joined into a cluster before $b_{n}$ is added.

\textbf{Third step: rows $k+1$ to $n$.} Here is the situation after row $k$: We have the same clusters in $S^+$ as after $k$ steps in the smaller data set $S$, except that the last cluster (the one which contains $a_{n}$) additionally contains the point $b_{n}$. In a diagram:
\[
 \xy
 \everyentry={\mathstrut}
 \xymatrix@C=3pt{
 a_0 & a_1 & a_2 & a_3 & a_4 & a_5 &\ldots & \ldots & a_m & \ldots & a_{n} & b_{n}
 }
 \POS
  "1,1"."1,3" *\frm<4pt>{-}
  ,"1,4"  *\frm<4pt>{-}
  ,"1,5"."1,6" *\frm<4pt>{-}
  ,"1,8"."1,12" *\frm<4pt>{-}
 \endxy
\]

The inter-cluster distances in $S^+$ from the cluster with $b_{n}$ to the other clusters might be smaller than without the point $b_{n}$  in $S^-$. We show, however, that this does not affect the remaining clustering steps:

In each step $r>k$, we have the following situation for some $x\leq y\leq s(k)$. The point $b_n$ might or might not be in the same cluster as $b_{s(r)}$.
\[
 \xy
 \everyentry={\mathstrut\vphantom{()_{()}}}
 \xymatrix@C=3pt{
 \ldots & a_x & \ldots & a_y& \ldots & a_{s(r)} & b_{s(r)}=a_{s(r)+1} & \ldots & \ldots & b_{n-1}
 }
 \POS
  "1,2"."1,6" *\frm<4pt>{-}
  , "1,8"+U
  \turnradius{4pt}%
  \PATH ~={**\dir{-}}
  `l "1,7"+L
  ` "1,7"+DR
  "1,8"+D
  \POS
  , "1,9"+U
  \PATH ~={**\dir{-}}
  `r "1,10"+R
  ` "1,9"+D
  "1,9"+D
 \endxy
\]
Let the distance $\delta_{s(r)}$ be realized as the distance from $b_{s(r)}$ to $a_y$. From Observation (a) and \autoref{76:argmin} in \nameref{alg:MST-linkage-core}, we know that this distance is minimal among the distances from $X\coloneq\{a_0,\ldots, a_{s(r)}\}$ to all other points in $S_0\setminus X$. In particular, the distance from $X$ to $b_{n}\in S_0\setminus X$ is not smaller than $\delta_{s(r)}$.

This proves that the addition of $b_{n}$ in step $k$ does not change the single linkage clustering in any later step $r>k$. This completes the inductive proof of \autoref{stronger}
\end{proof}

\section{Performance}\label{sec:performance}

In this section, we compare the performance of the algorithms and give recommendations on which algorithm to choose for which clustering method. We compare both the theoretical, asymptotic worst-case performance, and the use-case performance on a range of synthetic random data sets.

\subsection{Asymptotic worst-case performance}\label{sec:worstcase}

Let $N$ denote the problem size, which is in this case the number of input data points. The input size is $\binom N2\in\Theta(N^2)$.

The asymptotic run-time complexity of \nameref{alg:MST-linkage-core} is obviously $\Theta(N^2)$, since there are two nested levels of loops in the algorithm (\autoref{MST:loop1} and implicitly \autoref{MST:loop2}). The run-time complexity of the \nameref{alg:NN-chain-core} algorithm is also $\Theta(N^2)$ \citep[page 86]{Murtagh1985}. Postprocessing is the same for both algorithms and is less complex, namely $O(N\log N)$ for sorting and $\Theta(N)$ for \nameref{convert}, so the overall complexity is $\Theta(N^2)$. This is optimal (in the asymptotic sense): the lower bound is also quadratic since all $\Theta(N^2)$ input values must be processed.

The NN-chain algorithm needs a writable working copy of the input array to store intermediate dissimilarities and otherwise only $\Theta(N)$ additional memory.

The generic algorithm has a best-case time complexity of $\Theta(N^2)$, but without deeper analysis, the worst-case complexity is $O(N^3)$. The bottleneck is \autoref{line:bottleneck} in \nameref{alg:generic_linkage}: In $O(N)$ iterations, this line might be executed up to $O(N)$ times and does a minimum search over $O(N)$ elements, which gives a total upper bound of $O(N^3)$. This applies for all clustering schemes except single linkage, where the loop starting at \autoref{while:start} is never executed and thus the worst-case performance is $\Theta(N^2)$. The memory requirements for the generic algorithm are similar to the NN-chain algorithm: a working copy of the dissimilarity array and additionally only $\Theta(N)$ temporary memory.

In contrast, the MST algorithm does not write to the input array $d$. All other temporary variables are of size $O(N)$. Hence, \nameref{alg:MST-linkage} requires no working copy of the input array and hence only half as much memory as \nameref{alg:generic_linkage} and \nameref{alg:NN-chain} asymptotically.

Anderberg's algorithm \citep[pages 135--136]{Anderberg1973} has the same asymptotic bounds as our generic algorithm. The performance bottleneck are again the repeated minimum searches among the updated dissimilarities. Since the generic algorithm defers minimum searches to a later point in the algorithm (if they need to be performed at all, by then), there are at least as many minimum searches among at least as many elements in Anderberg's algorithm as in the generic algorithm. The only point where the generic algorithm could be slower is the maintenance of the priority queue with nearest neighbor candidates, since this does not exist in Anderberg's algorithm. The bottleneck here are potentially up to $O(N^2)$ updates of a queue in \autoref{line:update queue} of \nameref{alg:generic_linkage}. In the implementation in the next section, the queue is realized by a binary heap, so an update takes $O(\log N)$ time. This could potentially amount to $O(N^2\log N)$ operations for maintenance of the priority queue. However, a reasonable estimate is that the saved minimum searches in most cases save more time than the maintenance of the queue with $O(N)$ elements costs, and hence there is a good reason to believe that the generic algorithm is at least as fast as Anderberg's algorithm.

Note that the maintenance effort of the priority queue can be easily reduced to $O(N^2)$ instead of $O(N^2\log N)$ worst case:
\begin{itemize}
\item A different priority queue structure can be chosen, where the ``decrease-key'' operation takes only $O(1)$ time. (Note that the bottleneck operation in \autoref{line:update queue} of \nameref{alg:generic_linkage} never increases the nearest-neighbor distance, only decreases it.) The author did not test a different structure since a binary heap convinces by its simple implementation.
\item Changed keys (minimal distances) need not be updated in the priority queue immediately. Instead, the entire queue might be resorted/regenerated at the beginning of every iteration. This takes $N-1$ times $O(N)$ time with a binary heap. Although this lowers the theoretical complexity for the maintenance of the binary queue, it effectively slowed down the algorithms in practice by a small margin. The reason is, of course, that the number and complexity of updates of the priority queue did by far not reach their theoretical upper bound in our test data sets (see below). Altogether, the maintenance of the priority queue, as proposed in \autoref{fig:generic_linkage} seems quite optimal from the practical perspective.
\end{itemize}

\subsection{Use-case performance}\label{sec:use-case}

In addition to the theoretical, asymptotic and worst-case considerations, we also measured the practical performance of the algorithms. \autoref{fig:performance} shows the run-time of the algorithms for a number of synthetic test data sets (for details see below). The solid lines are the average over the data sets. (The graphs labeled ``Day-Edelsbrunner'' are discussed in \autoref{sec:alternatives}.) The lightly colored bands show the range from minimum to maximum time over all data sets for a given number of points.

\setcounter{figure}{\arabic{algorithm}}
\begin{figure}
\begin{tabular*}\textwidth{@{}l@{\extracolsep{\fill}}c@{}}
Method: ``single'' &$\vcenter{\hbox{\includegraphics{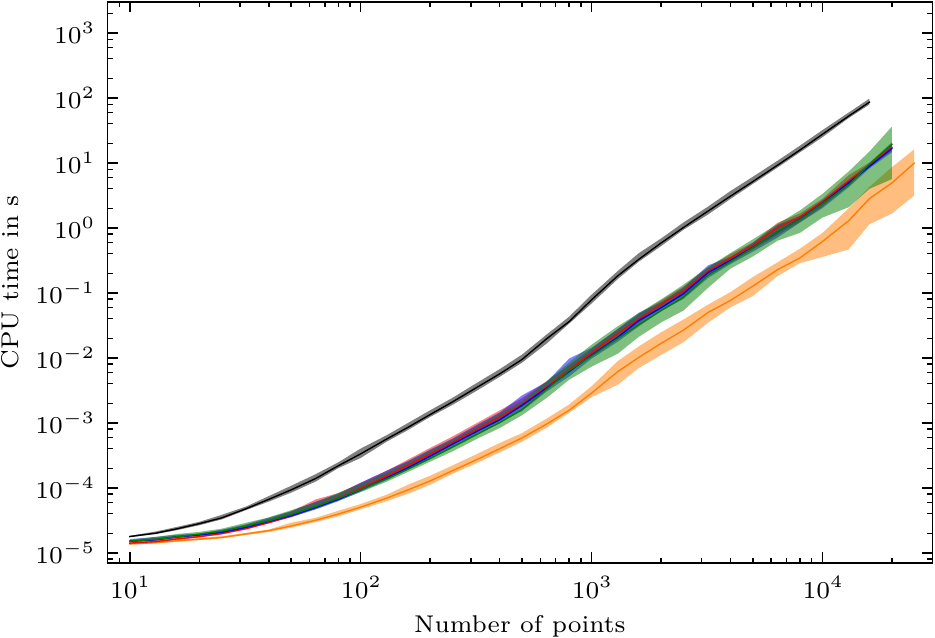}}}$\\
\parbox{1.8in}{\raggedright Method: ``average''\\[6pt]``complete'', ``weighted'' and ``Ward'' look very similar.} &$\vcenter{\hbox{\includegraphics{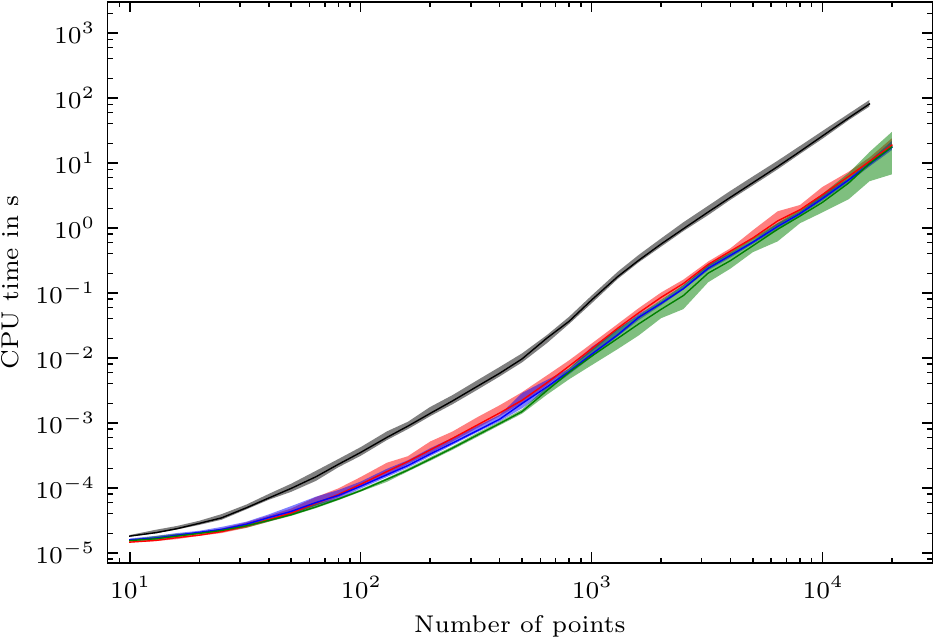}}}$\\
\parbox{1.8in}{\raggedright Method: ``centroid''\\[6pt]``median'' looks very similar.} &$\vcenter{\hbox{\includegraphics{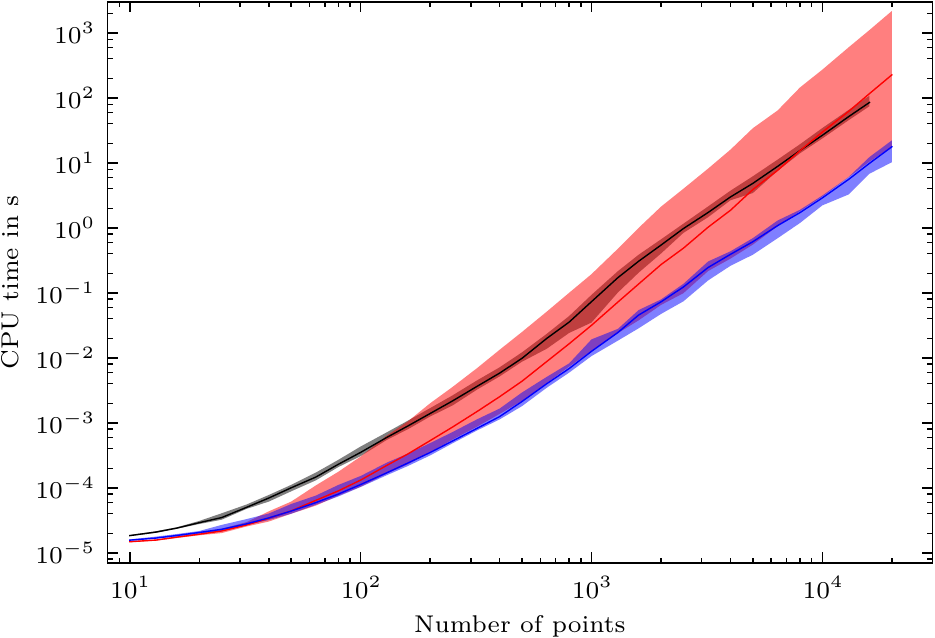}}}$\\
\end{tabular*}
\vskip-4pt
\newcommand{\sq}{{\setbox0\hbox{E}\vrule width\wd0 height\wd0 depth0pt\relax}}
\caption{Performance of several SAHN clustering algorithms. Legend: \textcolor{blue}{\sq}~Generic algorithm (\autoref{fig:generic_linkage}), \textcolor{red}{\sq}~\citet[pages 135--136]{Anderberg1973}, \textcolor{green!70!black}{\sq}~NN-chain algorithm (\autoref{fig:NN-chain}), \textcolor{orange}{\sq}~MST-algorithm (\autoref{fig:SL}), \textcolor{black}{\sq}~\citet[Table 5]{Day_Edelsbrunner1984}.}\label{fig:performance}
\end{figure}
\setcounter{algorithm}{\arabic{figure}}

The following observations can be made:
\begin{itemize}
\item For single linkage clustering, the MST-algorithm is clearly the fastest one. Together with the fact that it has only half the memory requirements of the other algorithms (if the input array is to be preserved), and thus allows the processing of larger data sets, the MST-algorithm is clearly the best choice for single linkage clustering.

\item For the clustering schemes without inversions (all except ``centroid'' and ``median''), the generic algorithm, the NN-chain algorithm and Anderberg's  algorithm have very similar performance.

The NN-chain algorithm is the only one with guaranteed $O(N^2)$ performance here. We can conclude that the good worst-case performance can be had here without any cut-backs to the use-case performance.

\item For the ``centroid'' and ``median'' methods, we see a very clear disadvantage to Anderberg's algorithm. Here, the worst case cubic time complexity occurs already in the random test data sets. This happens with great regularity, over the full range of input sizes. Our \nameref{alg:generic_linkage} algorithm, on the other hand, does not suffer from this weakness: Even though the theoretical worst-case bounds are the same, the complexity does not raise above the quadratic behavior in our range of test data sets. Hence, we have grounds to assume that  \nameref{alg:generic_linkage} is much faster in practice.
\end{itemize}

\subsection{Conclusions}

Based on the theoretical considerations and use-case tests, we can therefore recommend algorithms for the various distance update schemes as follows:

\begin{itemize}
\item ``single'' linkage clustering: The MST-algorithm is the best, with respect to worst-case complexity, use-case performance and memory requirements.

\item ``complete'', ``average'', ``weighted'', ``ward'': The NN-chain algorithm is preferred, since it guarantees $O(N^2)$ worst case complexity without any disadvantage to practical performance and memory requirements.

\item ``centroid'', ``median'': The generic clustering algorithm is the best choice, since it can handle inversions in the dendrogram and the performance exhibits quadratic complexity in all observed cases.
\end{itemize}

Of course, the timings in the use-case tests depend on implementation, compiler optimizations, machine architecture and the choice of data sets. Nevertheless, the differences between the algorithms are very clear here, and the comparison was performed with careful implementations in the identical environment.

The test setup was as follows: All algorithms were implemented in C++ with an interface to Python \citep{Python}\ and the scientific computing package NumPy \citep{NumPy} to handle the input and output of arrays. The test data sets are samples from mixtures of multivariate Gaussian distributions with unity covariance matrix in various dimensions (2, 3, 10, 200) with various numbers of modes (1, 5, $[\sqrt N]$), ranging from $N=10$ upwards until memory was exhausted ($N=20000$ except for single linkage). The centers of the Gaussian distributions are also distributed by a Gaussian distribution. Moreover, for the methods for which it makes sense (single, complete, average, weighted: the ``combinatorial'' methods), we also generated $10$ test sets per number of input points with a uniform distribution of dissimilarities.

The timings were obtained on a PC with an Intel dual-core CPU T7500 with 2.2 GHz clock speed and 4GB of RAM and no swap space. The operating system was Ubuntu 11.04 64-bit, Python version: 2.7.1, NumPy version: 1.5.1, compiler: GNU C++ compiler, version 4.5.2. Only one core of the two available CPU cores was used in all computations.

\section{Alternative algorithms}\label{sec:alternatives}

The MST algorithm has the key features that it (1) needs no working copy of the $\Theta(N^2)$ input array and only $\Theta(N)$ working memory, (2) is fast since it reads every input dissimilarity only once and otherwise deals only with $\Theta(N)$ memory. There is a second algorithm with these characteristics, Sibson's SLINK algorithm \citep{Sibson1973}. It is based on the insight that a single linkage dendrogram for $N+1$ points can be computed from the dendrogram of the first $N$ points plus a single row of distances $(d[N,0],\ldots, d[N,N-1])$. In this fashion, the SLINK algorithm even reads the input dissimilarities in a fixed order, which can be an advantage over the MST algorithm if the favorable input order can be realized in an application, or if dissimilarities do not fit into random-access memory and are read from disk.

\looseness 1 However, there is one important difference: even though the output data format looks deceptively similar to the MST algorithm (the output can be converted to a stepwise dendrogram by exactly the same process: sorting with respect to dissimilarities and a union-find procedure to generate node labels from cluster representatives), the SLINK algorithm cannot handle ties. This is definite, since e.g.\ the output in the example situation on page \pageref{tie data set} is the same in all three cases, and hence no postprocessing can recover the different stepwise dendrograms.

There is an easy way out by specifying a secondary order
\[
  d(i,j) \prec d(k,l) \quad\mathrel{\mathpalette\centercolon\Longleftrightarrow}\quad
\begin{cases}
 d(i,j) < d(k,l)&\text{if this holds,}\\
 Ni+j < Nk+l& \text{if }d(i,j)=d(k,l)
\end{cases}
\]
to make all dissimilarities artificially distinct. In terms of performance, the extra comparisons put a slight disadvantage on the SLINK algorithm, according to the author's experiments. However, the difference is not much, and the effect on timings may be compensated or even reversed in a different software environment or when the input order of dissimilarities is in favor of SLINK. Hence, the SLINK algorithm is a perfectly fine tool, as long as care is taken to make all dissimilarities unique.

The same idea of generating a dendrogram inductively is the basis of an algorithm by \citet{Defays1977}. This paper is mostly cited as a fast algorithm for complete linkage clustering. However, it definitely is not an algorithm for complete linkage clustering, as the complete linkage method is commonly defined, in this paper and identically elsewhere.

An algorithm which is interesting from the theoretical point of view is given by \citet[Table 5]{Day_Edelsbrunner1984}. It uses $N$ priority queues for the nearest neighbor of each point. By doing so, the authors achieve a worst-case time complexity of $O(N^2\log N)$, which is better than the existing bound $O(N^3)$ for the schemes where the NN-chain algorithm cannot be applied. The overhead for maintaining a priority queue for each point, however, slows the algorithm down in practice. The performance measurements in \autoref{fig:performance} include the Day-Edelsbrunner algorithm. Day and Edelsbrunner write their algorithm in general terms, for any choice of priority queue structure. We implemented the algorithm for the measurements in this paper with binary heaps, since these have a fixed structure and thus require the least additional memory. But even so, the priority queues need additional memory of order $\Theta(N^2)$ for their bookkeeping, which can also be seen in the graphs since they stop at fewer points, within the given memory size of the test. The graphs show that even if the Day-Edelsbrunner algorithm gives the currently best asymptotic worst-case bound for the ``centroid'' and ``median'' methods, it is inefficient for practical purposes.

\citet[\textsection\,II]{Krivanek} suggested to put all $\binom N2$ dissimilarity values into an $(a,b)$-tree data structure. He claims that this enables hierarchical clustering to be implemented in $O(N^2)$ time. Křivánek's conceptually very simple algorithm relies on the fact that $m$ insertions into an $(a,b)$-tree can be done in $O(m)$ amortized time. This is only true when the positions, where the elements should be inserted into the tree, are known. Searching for these positions takes $O(\log N)$ time per element, however. (See \citet[\textsection\,2.1.2]{Mehlhorn_Tsakalidis} for an accessible discussion of amortized complexity for $(2,4)$-trees; \citet{Huddleston_Mehlhorn} introduce and discuss $(a,b)$-trees in general.) Křivánek did not give any details of his analysis, but based on his short remarks, the author cannot see how Křivánek's algorithm achieves $O(N^2)$ worst-case performance for SAHN clustering.

\section{Extension to vector data}\label{sec:vectordata}

If the input to a SAHN clustering algorithm is not the array of pairwise dissimilarities but $N$ points in a $D$-dimensional real vector space, the lower bound $\Omega(N^2)$ on time complexity does not hold any more. Since much of the time in an SAHN clustering scheme is spent on nearest-neighbor searches, algorithms and data structures for fast nearest-neighbor searches can potentially be useful. The situation is not trivial, however, since (1) in the ``combinatorial'' methods (e.g.\ single, complete, average, weighted linkage) the inter-cluster distances are not simply defined as distances between special points like cluster centers, and (2) even in the ``geometric'' methods (the Ward, centroid and median schemes), points are removed and new centers added with the same frequency as pairs of closest points are searched, so a \emph{dynamic} nearest-neighbor algorithm is needed, which handles the removal and insertion of points efficiently.

Moreover, all known fast nearest-neighbor algorithms lose their advantage over exhaustive search with increasing dimensionality. Additionally, algorithms will likely work for one metric in $\mathbb R^D$ but not universally. Since this paper is concerned with the general situation, we do not go further into the analysis of the ``stored data approach'' \citep[\textsection\,6.3]{Anderberg1973}. We only list at this point what can be achieved with the algorithms from this paper. This will likely be the best solution for high-dimensional data or general-purpose algorithms, but there are better solutions for low-dimensional data outside the scope of this paper. The suggestions below are at least helpful to process large data sets since memory requirements are of class $\Theta(ND)$, but they do not overcome their $\Omega(N^2)$ lower bound on time complexity.
\begin{itemize}
\item The MST algorithm for single linkage can compute distances on-the-fly. Since every pairwise dissimilarity is read in only once, there is no performance penalty compared to first computing the whole dissimilarity matrix and then applying the MST algorithm. Quite the contrary, computing pairwise distances in-process can result in faster execution since much less memory must be reserved and accessed. The MST algorithm is suitable for any dissimilarity measure which can be computed from vector representations (that is, all scale types are possible, e.g.\ $\mathbb R$-valued measurements, binary sequences and categorical data).

\item The NN-chain algorithm is suitable for the ``Ward'' scheme, since inter-cluster distances can be defined by means of centroids as in \autoref{fig:formula}. The initial inter-point dissimilarities must be Euclidean distances (which is anyway the only setting in which Ward linkage describes a meaningful procedure).

\item The generic algorithm is suitable for the ``Ward'', ``centroid'' and ``median'' scheme on Euclidean data. There is a simpler variant of the \nameref{alg:generic_linkage} algorithm in \autoref{fig:variant}, which works even faster in this setting.
\begin{algorithm}\label{fig:variant}
\begin{algorithmic}[1]
\Procedure{Generic\_\hskip0pt linkage\_\hskip0pt variant}{$N, d$}\label{alg:generic_linkage_variant}
\Statex{$d$ is either an array or a function which computes dissimilarities from cluster centers.}
\Statex \hskip\algorithmicindent$\vdots$
\Statex
(Lines 2 to 13 are the same as in \nameref{alg:generic_linkage}.)
\makeatletter
\setcounter{ALG@line}{4}
\makeatother
\For{$x$ in $S\setminus\{N-1\}$}
\Statex \hskip2\dimexpr\algorithmicindent\relax$\vdots$
\makeatletter
\setcounter{ALG@line}{13}
\makeatother
\While{$b\notin S$}
\Statex \hskip3\dimexpr\algorithmicindent\relax $\vdots$
\Comment{Recalculation of nearest neighbors, if necessary.}
\makeatletter
\setcounter{ALG@line}{20}
\makeatother
 \EndWhile
 \State{Remove $a$ and $b$ from $Q$.}
 \State{Append $(a,b,\delta)$ to $L$.}
 \State{Create a new label $n\gets -i$}
 \State{$\size[n]\gets\size[a]+\size[b]$}
 \State{$S\gets (S\setminus\{a,b\})\cup \{n\}$}
\Statex
 \For{$x$ in $S\setminus\{n\}$}
 \Comment{Extend the distance information.}
   \State{$d[x,n]\gets d[n,x] \gets \nameref{formula}(d[a,x], d[b,x], d[a,b], \size[a], \size[b], \size[x])$}
 \EndFor
\Statex {or}
\makeatletter
\setcounter{ALG@line}{26}
\makeatother
\State{Compute the cluster center for $n$ as in \autoref{fig:formula}.}
\Statex
\makeatletter
\setcounter{ALG@line}{29}
\makeatother
 \State{$\nnghbr[n]\gets\argmin_{x>n} d[n,x]$}
 \State{Insert $(n, d[n,\nnghbr[n]])$ into $\mindist$ and $Q$}
 \EndFor
 \State{\Return{$L$}}
\EndProcedure
\end{algorithmic}
\caption{The generic clustering algorithm (variant).}
\end{algorithm}
The principle of the algorithm \nameref{alg:generic_linkage_variant} is the same: each array entry $\mindist[x]$ maintains a lower bound on all dissimilarities $d[x,y]$ for nodes with label $y>x$. The \nameref{alg:generic_linkage} algorithm is designed to work efficiently with a large array of pairwise dissimilarities. For this purpose, the join of two nodes $a$ and $b$ re-uses the label $b$, which facilitates in-place updating of the dissimilarity array in an implementation. The \nameref{alg:generic_linkage_variant} algorithm, in contrast, generates a unique new label for each new node, which is smaller than all existing labels. Since the new label is at the beginning of the (ordered) list of nodes and not somewhere in the middle, the bookkeeping of nearest neighbor candidates and minimal distances is simpler in \nameref{alg:generic_linkage_variant}: in particular, the two loops in lines \ref{dispose:start}--\ref{dispose:end} of \nameref{alg:generic_linkage} can be disposed of entirely. Moreover, experiments show that \nameref{alg:generic_linkage_variant} needs much less recalculations of nearest neighbors in some data sets. However, both algorithms are similar, and which one is faster in an implementation seems to depend strongly on the actual data structures and their memory layout.
\end{itemize}

Another issue which is not in the focus of this paper is that of parallel algorithms. For the ``stored matrix approach'', this has a good reason since the balance of memory requirements versus computational complexity does not make it seem worthwhile to attempt parallelization with current hardware. This changes for vector data, when the available memory is not the limiting factor and the run-time is pushed up by bigger data sets. In high-dimensional vector spaces, the advanced clustering algorithms in this paper require little time compared to the computation of inter-cluster distances. Hence, parallelizing the nearest-neighbor searches with their inherent distance computations appears a fruitful and easy way of sharing the workload. The situation becomes less clear for low-dimensional data, however.

\section{Conclusion}\label{sec:conclusion}

Among the algorithms for sequential, agglomerative, hierarchic, nonoverlapping (SAHN) clustering on data with a dissimilarity index, three current algorithms are most efficient: Rohlf's algorithm \nameref{alg:MST-linkage} for single linkage clustering, Murtagh's algorithm \nameref{alg:NN-chain} for the ``complete'', ``average'', ``weighted'' and ``Ward'' schemes, and the author's \nameref{alg:generic_linkage} algorithm for the ``centroid'' and ``median'' schemes and the ``flexible'' family. The last algorithm can also be used for an arbitrary distance update formula. There is even a simpler variant \nameref{alg:generic_linkage_variant}, which seems to require less internal calculations, while the original algorithm is optimized for in-place updating of a dissimilarity array as input. The \nameref{alg:generic_linkage} algorithm and its variant are new; the other two algorithms were described before, but for the first time they are proved to be correct.

\section*{Acknowledgments}
This work was funded by the National Science Foundation grant DMS-0905823 and the Air Force Office of Scientific Research grant FA9550-09-1-0643.

\bibliography{Clustering_algorithms_arXiv}
\vspace{1cm}

\parindent0pt\parskip\medskipamount\raggedright
\textsc{Daniel Müllner\\Stanford University\\Department of Mathematics\\450 Serra Mall, Building 380\\Stanford, CA 94305}\par
E-mail: \href{mailto:muellner@math.stanford.edu}{\texttt{muellner@math.stanford.edu}}\par
\url{http://math.stanford.edu/~muellner}
\end{document}